\documentclass[runningheads]{llncs}

\PassOptionsToPackage{numbers, sort&compress}{natbib}

\usepackage[dvipsnames]{xcolor}
\usepackage{graphicx}
\usepackage{adjustbox}
\usepackage{relsize}
\usepackage{bigstrut}
\usepackage{tikz}
\usetikzlibrary{backgrounds}
\usetikzlibrary{arrows,shapes}
\usetikzlibrary{tikzmark}
\usetikzlibrary{calc}

\usepackage{comment}
\usepackage{amssymb,amsmath} %
\usepackage{color}

\usepackage[accsupp]{axessibility}  %

\colorlet{mhpurple}{Plum!80}
\usepackage{varwidth}
\usepackage{capt-of}
\usepackage{multirow,booktabs}
\usepackage{makecell}
\usepackage{tabularx}
\usepackage{xspace}
\usepackage[capitalize]{cleveref}
\usepackage{textcomp}
\usepackage{mathtools, nccmath}
\usepackage{wrapfig}
\usepackage{comment}
\usepackage{array}
\usepackage{ragged2e}
\newcolumntype{P}[1]{>{\RaggedRight\hspace{0pt}}p{#1}}
\newcolumntype{X}[1]{>{\RaggedRight\hspace*{0pt}}p{#1}}

\usepackage{amsmath,amsfonts,bm}

\def\eqref#1{equation~\ref{#1}}

\def\1{\bm{1}}

\newcommand{\test}{\mathcal{D_{\mathrm{test}}}}

\def\rvb{{\mathbf{b}}}

\def\rvw{{\mathbf{w}}}
\def\rvx{{\mathbf{x}}}

\def\rvz{{\mathbf{z}}}

\def\ervo{{\textnormal{o}}}

\def\rmA{{\mathbf{A}}}

\def\rmI{{\mathbf{I}}}

\def\rmQ{{\mathbf{Q}}}
\def\rmR{{\mathbf{R}}}

\def\rmU{{\mathbf{U}}}
\def\rmV{{\mathbf{V}}}
\def\rmW{{\mathbf{W}}}

\def\rmY{{\mathbf{Y}}}
\def\rmZ{{\mathbf{Z}}}

\DeclareMathAlphabet{\mathsfit}{\encodingdefault}{\sfdefault}{m}{sl}
\SetMathAlphabet{\mathsfit}{bold}{\encodingdefault}{\sfdefault}{bx}{n}

\DeclareMathOperator*{\argmin}{arg\,min}

\DeclareRobustCommand*{\plus}{\raisebox{-0.1\height}{\scalebox{1.05}{\texttt{+}}}}

\newcommand{\bftab}{\fontseries{b}\selectfont}
\newcommand{\etal}{\textit{et al.}}
\newcommand{\ie}{\textit{i.e.}}
\newcommand{\eg}{\textit{e.g.}}
\newcommand{\method}{SphereFed\xspace}
\newcommand{\methodd}{FFC\xspace}
\newcommand{\noniid}{non-\textit{i.i.d.} }
\newcommand{\Noniid}{Non-\textit{i.i.d.} }
\newcommand{\iid}{\textit{i.i.d.} }

\newcommand{\fakeparagraph}[1]{\vspace{0.618em}\noindent\textbf{#1}}

\crefname{section}{Sec.}{Secs.}
\Crefname{section}{Section}{Sections}
\Crefname{table}{Table}{Tables}
\crefname{table}{Tab.}{Tabs.}

\definecolor{draw_org}{HTML}{f4b247}
\definecolor{draw_green}{HTML}{66c56c}

\begin{document}
\pagestyle{headings}
\mainmatter
\def\ECCVSubNumber{2255}  %

\title{SphereFed: Hyperspherical Federated Learning} %

\titlerunning{SphereFed: Hyperspherical Federated Learning}
\author{Xin Dong\inst{1} \and
Sai Qian Zhang\inst{1} \and
Ang Li\inst{2} \and H.T. Kung\inst{1}}
\authorrunning{X. Dong et al.}
\institute{Harvard University \and UT Dallas\\
\textsf{\small xindong@g.harvard.edu}}
\maketitle

\vspace{-6mm}
\begin{abstract}
Federated Learning aims at training a global model from multiple decentralized devices (\textit{i.e.} clients) without exchanging their private local data. A key challenge is the handling of \noniid (independent identically distributed)  data across multiple clients that 
may induce disparities of their local features. %
We introduce the Hyperspherical Federated Learning (\method) framework to address the \noniid issue by constraining learned representations of data points to be on a unit hypersphere shared by clients. 
Specifically, all clients learn their local representations by minimizing the loss with respect to a fixed classifier 
whose weights span the unit hypersphere.
After federated training in improving the global model,  this classifier is further calibrated with a closed-form solution by minimizing a mean squared loss. 
We show that 
the calibration solution can be computed efficiently 
and distributedly without 
direct access of local data. Extensive experiments indicate that our \method approach is able to improve the accuracy of multiple existing federated learning algorithms by a considerable margin (up to 6\% on challenging datasets) with enhanced computation and communication efficiency across datasets and model architectures.

\end{abstract}

\section{Introduction}
Federated learning (FL) is an emerging machine learning paradigm in which distributed clients learn on private data and communicate with a coordinating server to train a single global model that generalizes well across local data~\cite{mcmahan2017communication,smith2017federated}. One of its major challenges is the the handling of \noniid (independent identically distributed) local data across clients~\cite{karimireddy2020scaffold,li2021federated,li2020federateds}. \Noniid local data leads to disparity of local models after learning on private data~\cite{zhao2018federated}. For instance, different feature\footnote{The terms representation and feature are used interchangeably.} extractors in local models may learn biased and discrepant input-to-feature mapping functions for the same class~\cite{duan2021feddna,li2021model,zhang2021federated}. This obstructs the convergence of collaborative training.

Existing federated learning algorithms primarily tackle the \noniid problem in two phases: (\romannumeral1) \textit{In the local learning phase}, regularization terms~\cite{acar2021federated,li2020federated,xu2021acceleration} and additive objectives~\cite{li2021model,yoon2021federated,zhu2021data} are used to control distances among local models via constraining the learning process. (\romannumeral2) \textit{In the post-learning phase}, the inevitable divergence of local models is corrected with additional information exchange~\cite{hsu2019measuring,karimireddy2020scaffold,shoham2019overcoming,xu2022coordinating,yao2020continual} and advanced aggregation strategies such as normalized averaging~\cite{wang2020tackling}, distillation~\cite{li2019fedmd}, and so on~\cite{acar2021debiasing,alshedivat2021federated,reddi2020adaptive,yuan2021federated}. 

In this work, we argue that federated learning can also be improved \textit{in the pre-learning phase}; this is a novel research direction complementary to existing approaches. A key insight is the use of a fixed classifier (\eg, the last fully-connected layer) that serves as a template of the feature extractor's output for all clients. Note the loss function is often computed on the inner product between the feature vector (\ie, output of the feature extractor) and the classifier's weight vectors. %
During local training, the feature extractor is optimized to project data from the $i$-th class to feature vectors that have the maximum inner product with the $i$-th of row of the classifier. We refer to the classifier as learning target of the feature extractor.
However, higher data heterogeneity leads to a larger disparity of classifiers (in terms of both norms and directions) across clients. In this regard, if the local classifiers can be aligned, clients would have more consistent learning targets without modifying the learning procedure. Unfortunately, a real-time classifier synchronization carries prohibitively high communication overhead for federated learning. To avoid this communication cost, we use instead, for all clients, a fixed classifier constructed from orthonormal basis vectors.

\begin{figure}[t]
    \centering
    \includegraphics[width=\linewidth]{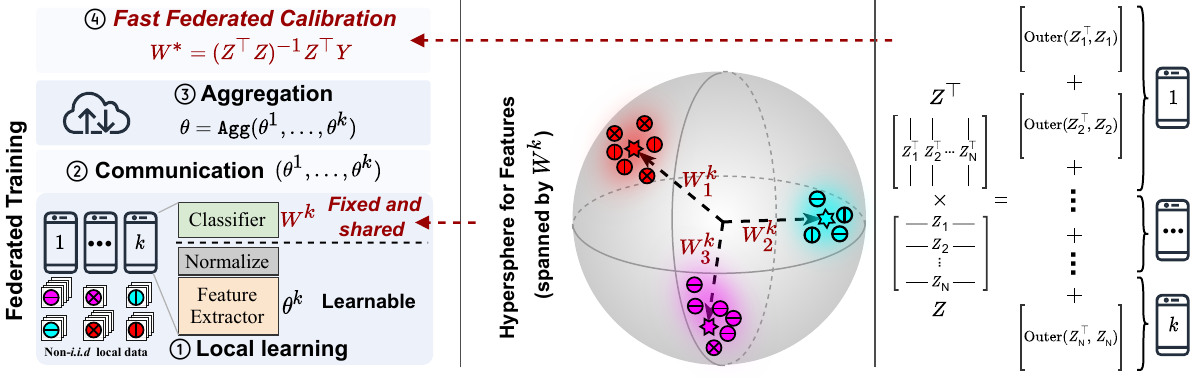}
    \vspace{-6mm}
    \caption{\textit{Left}: An overview of SphereFed (Hyperspherical Federated Learning). Before federated training starts, we construct a fixed shared classifier whose weight vectors span an unit hypersphere. After federated training ends, we calibrate the classifier in a distributed manner. \textit{Middle}: All clients share the same hypersphere and learn to map local samples (markers represent clients) from the same class (colors represent classes) to the same area on the hypersphere whose centroid (the pentagram) corresponds to a weight vector. \textit{Right}: Leveraging the linearity of the classifier, we derive its closed-form optimum which can be precisely computed by distributed clients. $Z^\top Z$ and $Z^\top Y$ are computed distributively because a matrix multiplication can be implemented as an accumulation of outer products and each outer product depends only on one client.}
    \label{fig:teaser}
    \vspace{-4mm}
\end{figure}

Motivated by this insight, we propose to construct a classifier whose weight vectors span an unit hypersphere \textit{before the federated training starts}. Throughout federated training, this classifier is fixed and shared by all clients. Meanwhile, we also normalize the feature representation to the same hypersphere. During local learning, clients' feature extractors learn to map data samples from the \textit{same} class to the \textit{same} area on the hypersphere whose centriod is the corresponding row vector of the classifier.
As a result, the learned local features for data belonging to the same class are better aligned and the interference among local models are reduced, leading to an improved accuracy of the global model. 

We name our approach \textit{Hyperspherical Federated Learning} (\method), which is a generic framework compatible with existing federated learning algorithms. An overview of the framework is illustrated in \cref{fig:teaser}. \method does \textbf{not} introduce extra hyper-parameters nor requiring additional computation. In fact, a pre-defined classifier eliminates the need of communication and brings improved efficiency to the system. Given that the classifiers are frozen during federated training, we propose to calibrate the classifier after federated training to achieve its optimum in a provable and lightweight manner. 
We first derive the closed-form optimum of the classifier leveraging its linearity and find that this closed-form solution can be precisely computed in a distributed manner without direct access to the private features (or data). 
We name this calibration method~\textit{Fast Federated Calibration} (FFC), which is provable and efficient compared with state-of-the-art methods (\eg, \cite{luo2021no}) that depend on synthetically generated virtual features.

We conduct extensive experiments to demonstrate that the proposed SphereFed method is compatible with and complementary to several existing federated learning algorithms, capable of introducing up to 6\% improvement on testing accuracy. Further experiments show that our proposed calibration achieves a performance gain comparable to the oracle fine-tuning with real features, verifying its theoretical optimality. 
A set of ablation studies are further presented to understand the efficacy of each design component in SphereFed.

\vspace{-2mm}
\section{Related Work}
\vspace{-2mm}
\subsubsection{Standard Federated Learning.} Federated learning (FL) was originally proposed by~\cite{mcmahan2017communication,smith2017federated}. 
To address the \noniid problem, works have been pursued in two directions: imposing additional constraints in the local learning phase~\cite{li2020federated,shamir2014communication,shoham2019overcoming,yoon2021federated} and conducting weight correction in the post learning phase~\cite{luo2021no,reddi2020adaptive,wang2020tackling,yuan2021federated}. 
There have also been studies that tackle the \noniid issue by 
augmenting the on-client and on-server data with public~\cite{li2019fedmd} or synthetic~\cite{yoon2021fedmix,zhu2021data} samples. 

\fakeparagraph{Personalized Federated Learning.} Personalized federated learning (pFL) differs from FL, by relaxing the setting of standard FL to allow each client to have its personalized local model via, e. g., additional local epochs after standard FL~\cite{bui2019federated,liang2020think}. In general, a personalized local model is more likely to obtain better accuracy on a local test set than the single global model, but the personalized local model could be more biased and less general to data from other sources~\cite{jiang2019improving,wang2019federated}. So, pFL and FL has different use focuses and application scenarios. Inspired from transfer learning~\cite{puigcerver2020scalable,you2021logme},
a line of pFL methods learns local private 
parameters for the classifier but uses a shared feature extractor~\cite{arivazhagan2019federated,collins2021exploiting,sun2021partialfed}. A concurrent and most related work is \scalebox{0.85}{FedBABU} which finds that fixing the classifier during collaborative learning is beneficial to the personalization process. Although this finding is consistent to our observations to some extent, our contribution is substantially different from \scalebox{0.85}{FedBABU}. First, we focus on FL while \scalebox{0.85}{FedBABU} focuses on pFL. Second, 
we ensure a stable performance gain resulting from an in-depth analysis on the benefit of fixing classifier. Third, we further propose a provable calibration method to improve the classifier after federated training. 

\fakeparagraph{Decoupling Layers for Federated Learning.} Dealing layers at different depth with varying strategies has demonstrated effectiveness for many tasks in centralized training~\cite{hoffer2018fix} like few-shot learning~\cite{shao2021mhfc,YeHZS2021Learning}, domain adaption~\cite{jain2011online,venkat2020your} and meta-learning~\cite{oh2020boil,raghu2019rapid}. Such layer decoupling studies could also benefit federated learning applications. For instance, parameters from different layers can be updated and synchronized with different frequency to save communication cost~\cite{chen2021communication,chen2019communication,diao2020heterofl}. \scalebox{1}{FedRecon}~\cite{singhal2021federated} splits a model into global/local parts and reconstructs the local part on clients in each round to improve privacy and efficiency. \scalebox{1}{FedUFO}~\cite{zhang2021federated} resorts an adversary module to reduce the divergence of feature extractors on clients. 
A most related work is \scalebox{1}{CCVR}~\cite{luo2021no} which also focuses on the classifier. \scalebox{1}{CCVR} conducts on-server calibration for the classifier by fine-tuning it with virtual features sampled from Gaussian distributions. This work uses fundamentally different methodologies for the classifier and has higher performance gains and less communication/computation costs against \scalebox{1}{CCVR}. 

\fakeparagraph{Hyperspherical Representation.} To the best of our knowledge, this is the first work introducing hyperspherical representation to address the \noniid challenge in FL. This combination is not trivial but motivated by analytical justifications and empirical supports as elaborated in the remaining sections. Hyperspherical representation has been widely adapted by studies on face recognition~\cite{liu2017sphereface,zheng2018ring}, long-tail recognition~\cite{kang2019decoupling}, regression~\cite{mettes2019hyperspherical}, metric learning~\cite{zhai2018classification,zhu2020spherical} and contrastive learning~\cite{khosla2020supervised} to 
enhance the discriminative power of features. In this work, under the context of FL, we use hyperspherical features with fixed targets to align the learning objectives and minimize cross-party interference. 

\vspace{-2mm}
\section{Federated Learning with \Noniid Clients}
\vspace{-1mm}
\subsection{Terminologies} 
We consider $K$ clients and a central server in a federated learning system. Each client $k\in[K]$ has a local and private dataset $\mathcal{D}^k$. We focus on the \noniid data setting where local datasets could have heterogeneous distributions~\cite{li2021federated}. The goal is to train a single global classification model collaboratively which performs well on the global test set. The loss function is represented using $\mathcal L(\cdot, \cdot)$.

For a single training example $\left(\rvx, y\right)$, let $\rvz = f_\theta(\rvx) \in \mathbb{R}^l$ denote the $l$-dimension feature vector given a feature extractor $f_\theta(\cdot)$ parameterized by $\theta$. The classifier $h_\rmW(\cdot)$ takes $\rvz$ as input and makes the final prediction after a linear transformation $\ervo = h_\rmW(\rvz) = \rmW\rvz+\rvb$ with a weight matrix $\rmW\in\mathbb{R}^{C\times l}$, where $C$ is the number of classes. For simplicity, we omit the bias term $\rvb$ in future equations.

\begin{figure}[t]
    \centering
    \includegraphics[width=0.8\linewidth]{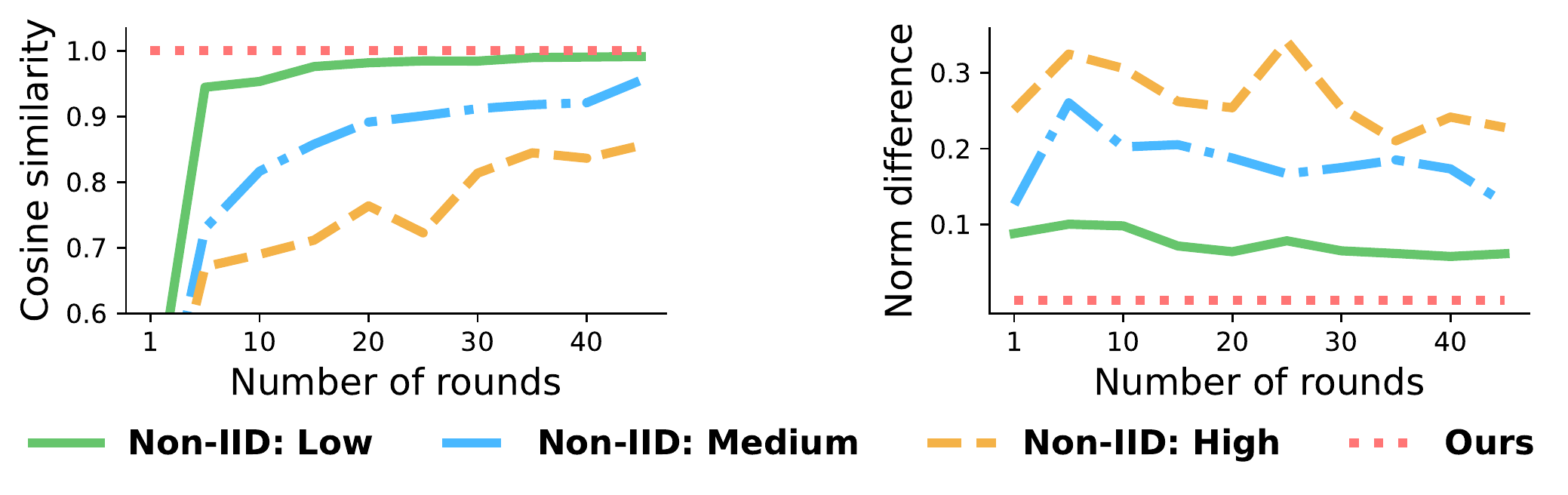}
    \vspace{-4mm}
    \caption{Direction and norm alignment of classifiers' weights across clients. For the sake of simplicity, we train ResNet18 on CIFAR-100 with 10 clients for 50 rounds, using FedAvg in this empirical study. The \noniid level is controlled by the concentration parameter of a Dirichlet distribution~\cite{karimireddy2020scaffold,li2021federated,li2020federateds}. There is a clear negative correlation between \noniid level and consistency of classifiers’ weights across clients, which incurs inconsistent local learning targets.}
    \vspace{-4mm}
    \label{fig:wei_consist}
\end{figure}

\subsection{\Noniid Data Leads to Inconsistent Local Learning Targets}
\label{sec:non_iid_lead_to_inconsis}

In each round of standard federated learning, each client optimizes the feature extractor and the classifier $(\theta, \rmW)$ jointly.
Then each client sends its updated feature extractor and classifier $(\theta^k, \rmW^k)$ to the central server which aggregates (\eg, averages~\cite{mcmahan2017communication}) all received local models into a single global one used for the next round. Prior studies~\cite{li2021federated} focus on either the local training loss function or an advanced aggregation strategy.

In this work, 
we pay special attention to the classifier.
A classifier is the closet layer to the loss function and the $i$-th row of its weights $\rvw_i^k$ acts as a feature template and the learning target of the $i$-th class for the feature extractor. 
As a result, the disparity of classifiers across clients induces inconsistent local learning targets and further engenders local feature extractors' disparity. A performance degradation may occur after aggregation in the central server.   

The above hypothesis is verified by empirical observations. To show that, we rewrite the output of the classifier as,
\begin{equation}
    \scalebox{0.9}{%
    $\rmW^k\rvz = \left[\rvw^k_1\rvz, \dots, \rvw^k_i\rvz,\dots, \rvw^k_C\rvz\right], \text{ where } \rvw^k_i\rvz = \|\rvw_i^k\|\|\rvz\|\cdot\cos{\left(\angle\left(\rvw_i^k,\rvz\right)\right)}$
    }.
    \label{equ:dotproduct2angle}
\end{equation}
$\angle(\cdot,\cdot)$ denotes the angle between two vectors and $\|\cdot\|$ is the euclidean norm of a vector. 
A local feature extractor $f_\theta^k$ learns to maximize the output of the ground-truth class, $\rvw_i^k\rvz,\ \forall i=y$, and minimize outputs of other classes $\rvw_j^k\rvz,\ \forall j\neq y$. In summary, \cref{equ:dotproduct2angle} highlights that both norms and directions of classifier's weight vectors impact the optimization of feature extractor and thus (at least partially) effect the distribution of features generated by the feature extractor. 

We empirically find that there is a clear negative correlation between \noniid degree and the consistency of classifiers' weights~(in terms of both norm and direction) across clients. We compute the cosine similarity (\cref{equ:cosine} and \cref{fig:wei_consist}, \textit{Left}) and norm difference (\cref{equ:norm_diff} and \cref{fig:wei_consist}, \textit{Right}) of classifier weight vectors for the same class but from arbitrary two different clients ($k_1\neq k_2$ and $1\leq k_1,k_2\leq K$) using FedAvg~\cite{mcmahan2017communication}.
\begin{center}
\vspace{-8mm}
\begin{tabular}{m{0.42\textwidth} m{0.55\textwidth}}
  \begin{equation}
  \scalebox{0.9}{%
    $\mathbb{E}_{c\sim [C], k_1\neq k_2}\left[\frac{\rvw_c^{k_1}\cdot\rvw_c^{k_2}}{\|\rvw_c^{k_1}\|\|\rvw_c^{k_2}\|}\right]$
    }
    \label{equ:cosine}
  \end{equation}
 &
  \begin{equation}
  \scalebox{0.9}{%
    $\mathbb{E}_{c\sim [C],k_1\neq k_2}\left[\left|\|\rvw_c^{k_{1}}\|-\|\rvw_c^{k_{2}}\|\right|\right]$
    }
  \label{equ:norm_diff}
  \end{equation}
\end{tabular}
\vspace{-6mm}
\end{center}
According to~\cref{fig:wei_consist}, data with a higher degree of \noniid is associated with less direction alignment and larger magnitude difference of classifier weight vectors among clients, which lead to weaker consistency of the local learning targets. This will further engender non-overlapped feature distributions for the same class on different clients as illustrated in~\cref{fig:feature_align_tsne} (\textit{Left}).

\section{Hyperspherical Federated Learning}
\label{sec:force}

\subsection{Hyperspherical Representation}
\label{sec:hyper_rep}

A simple yet effective tweak to bypass the aforementioned issue is to align features from clients on a hypersphere, aided by a fixed classifier.  In~\cref{equ:dotproduct2angle}, we show that both the norm and the direction of $k$-th weight vector $\rvw_i^k$ play crucial roles in the learning of $f_{\theta^k}$ but a \noniid data distribution causes a disorder of norms and directions. To tackle this problem, we consider to construct $\rmW^k=\{\rvw_i^k\}^C_{i=1}$ manually, which has a unit norm and orthogonal components,
\begin{equation}
    \scalebox{0.9}{%
    $\rmW^k = \left\{\rvw_i^k\right\}^C_{i=1}, \ \ \text{where}\ \  \|\rvw^k_i\|=1\ \text{and}\  \rvw^k_i \perp \rvw^k_j,\ \forall\, i=j$.
    }
\end{equation}
Note that $\smash{\{\rvw_i^k\}^C_{i=1}}$ span an $l$-dimension unit hypersphere. The orthogonality among $\smash{\{\rvw_i^k\}^C_{i=1}}$ ensures the maximum separation between arbitrary pair of classes. In addition, the uniformly unit norm guarantees balance in classes. 
Feature normalization is further adapted to project feature vectors on the same unit hypersphere, $\tilde{\rvz} = {\rvz}/{\|\rvz\|}$.
Normalizing features enables $f_\theta^k$ to focus on learning feature vectors' directions and makes federated training process more robust to feature magnitude. Given a data point $\left(\rvx, y\right)$, a feature extractor $f_{\theta^k}$ maps the input $\rvx$ to a feature vector $f_{\theta^k}(\rvx)/\|f_{\theta^k}(\rvx)\|$ on the unit hypersphere, using the corresponding weights $\rvw^k_y$ as the target of mapping. 

To ensure that all clients have the same learning targets (\ie, the same classifier $h_{\rmW^k}$), we share the constructed $\rmW^k$ with all clients and keep the shared $\rmW^k$ fixed throughout the federated training process. By doing this, local classifiers become consistent automatically without costly frequent inter-client synchronization. Since all clients now share the same learning targets, different local feature extractors on clients learn to map local data samples from the $i$-th class to the same area on the hypersphere with $\rvw_i^k$ as the centroid of that area. As a result, the norms and directions of local features are aligned with reduced interference across clients. In addition, the features from different classes have minimized overlaps and balanced magnitudes. An illustration of hyperspherical features can be found in~\cref{fig:teaser} (Middle).

\begin{figure}[t]
    \centering
    \includegraphics[width=0.9\textwidth]{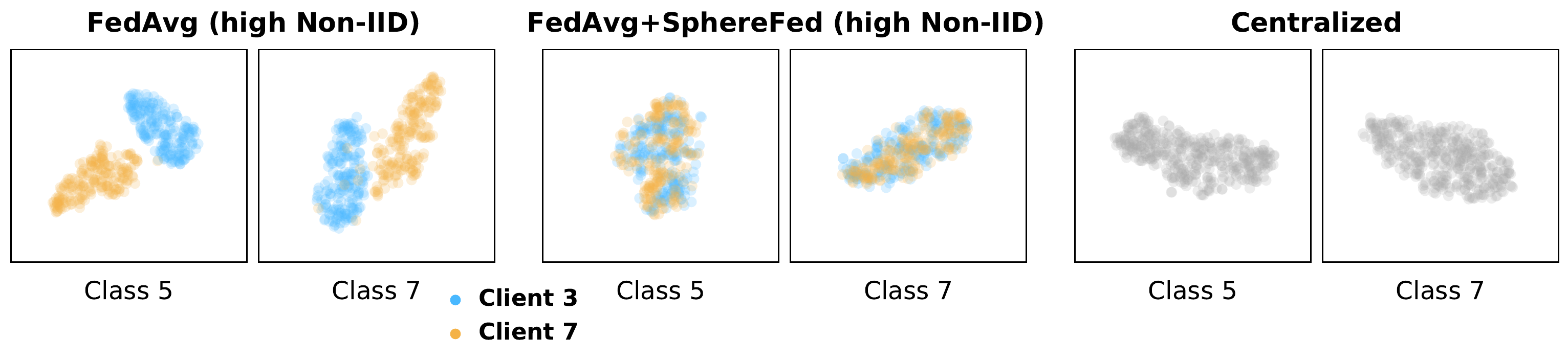}
    \vspace{-4mm}
    \caption{A qualitative study of Hyperspherical Federated Learning (\method). In the left and middle sub-figures, dots in different colors represent features of the same class but generated by local models $f_{\theta^k}$ of different clients. \method encourages consistency among clients' features by aligning local learning targets.}
    \label{fig:feature_align_tsne}
    \vspace{-4mm}
\end{figure}

The benefits of adapting the proposed hyperspherical features (and each design component described above) are revealed in a qualitative evaluation (\cref{fig:feature_align_tsne}). For more detailed ablation study including quantitative results, please refer to \cref{sec:exp:abla}. In~\cref{fig:feature_align_tsne}, we plot a certain class's features from different clients to visualize local features alignment across clients. Three kinds of methods are compared including FedAvg~\cite{mcmahan2017communication} (with conventional classifier), standard centralized training, and \method. We use MobileNetV2 for all three methods. For FedAvg and \method, we partition the CIFAR-100 dataset to 10 clients according to the Dirichlet distribution with the concentration parameter $\alpha$ set as $0.1$ to simulate the high \noniid scenario. For the sake of visualization, we randomly select two clients 
(\eg, the 3-rd and 7-th clients in~\cref{fig:feature_align_tsne})
and use their local models (at the $90$-th round) to generate features of two random classes 
(\eg, the class 5 and 7 in~\cref{fig:feature_align_tsne})
. For centralized training, we train MobileNetV2 for 120 epochs to convergence and use the learnt model to generate features. The dimension of raw features is 1280 and we use t-SNE~\cite{JMLR:v9:vandermaaten08a} to reduce the dimension to 2 for visualization. According to~\cref{fig:feature_align_tsne}, in FedAvg, local models learn divergent mapping functions and thus features from the same class but different clients are biased to different distributions (\ie, non-overlapped clusters in \cref{fig:feature_align_tsne}). While, our hyperspherical features aligns well across clients as the centralized training (\ie, fused clusters in \cref{fig:feature_align_tsne}).

\subsection{Using Mean Squared Error Loss on Hyperspherical Features}

 A widely accepted training method for classification tasks is to apply the softmax function~\cite{anzai2012pattern} to the classifier output $\rmW^k\rvz$ before calculating the cross entropy (CE) loss. Combination of CE and softmax function is known to be sensitive to the scale of its input~\cite{hoffer2018fix,vaswani2017attention}. However, in \method, the classifier's output, has less-than-one scale because of unit weights and hyperspherical features. 
 To mitigate such a scaling issue, prior work for centralized training adopts either a pre-defined~\cite{cheraghian2021semantic,vaswani2017attention} or a learnable~\cite{hoffer2018fix,kang2019decoupling} scaling parameter $\tau$ in $\tau\cdot\rmW^k\rvz$ (\textit{i.e.}, temperature) to stabilize optimization. Unfortunately, performing a grid search on the scaling hyper-parameter would add significant communication and computation overheads in the setting of federated learning. A freely learnable scaling parameter could also aggravate local models' disparity when the clients have different scaling parameters. 
 
 Interestingly, historical~\cite{golik2013cross,sangari2015convergence} and recent~\cite{achille2021lqf,hui2020evaluation,kornblith2021better} studies show competitive results of mean square error~(MSE)~\cite{brier1950verification} compared with CE for classification tasks on modern deep architectures. We refer readers to \cite{muthukumar2021classification} and the related literature~\cite{belkin2021fit,mai2019high,thrampoulidis2020theoretical} for a more in-depth theoretical discussion. 
In this work, we use MSE to learn hyperspherical features to bypass the scaling issue of CE, \ie,
\begin{equation}
    \label{equ:mse_loss}
    \scalebox{0.85}{%
    $\mathcal{L}_\textrm{{\tiny MSE}}(\rmW^k\rvz,\ y) =  \frac{1}{C}\|\rmW^k\rvz-\texttt{one\_hot}(y)\|^2 = \frac{1}{C}\sum_{i=1}^C\Big(\rvw^k_i\rvz-\bm{1}(i=y)\Big)^2,$}
\end{equation}
where $\bm{1}\small(i=y)$ is equal to 1 if and only if $i=y$, $C$ is the number of classes, and $\texttt{one\_hot}(\cdot)$ is the one-hot vector representation of a label.

\subsection{Fast Federated Calibration (FFC)}
\label{sec:mse_loss}
 Throughout federated training of the hyperspherical representation, the weight of classifier $\rmW^k$ is fixed to help align learning targets of local features. So naturally, after federated training, a calibration on the classifier 
could be useful in improving the accuracy of the resulting global model.
 We, in turn, fix the learnt global feature extractor and calibrate the global classifier to boost the performance of the global model.

An interesting effect of using MSE loss in conjunction with a linear classifier is that we are able to calculate the unique closed-form optimum of classifier's weight matrix given the input of the linear classifier. 
Formally, the objective of calibrating the classifier $\rmW$ is,
\begin{equation}
\label{equ:loss_for_w}
    \scalebox{0.85}{%
    $\argmin_\rmW\ \mathbb{E}_{(\rvx, y)\sim\mathcal{D}}\left[\mathcal{L}_\textrm{{\tiny MSE}}\left(\rmW\rvz,\ y\right)\right],\quad \textrm{where}\ \rvz = {f_\theta(\rvx)}/{\|f_\theta(\rvx)\|} \ \, \textrm{and}\ \, \mathcal{D} =  \bigcup_{k\in [K]}\mathcal{D}^k$
    }~.
\end{equation}
We temporarily remove the superscripts of $(\theta, \rmW)$ to emphasize that we consider the global feature extractor and global classifier. We refer to $\mathcal{D}$ as the whole dataset which consists of all local training data. 
\cref{equ:loss_for_w} is essentially a least square problem, which has a closed-form solution, \ie,
\begin{equation}
\label{equ:closed_form}
    \scalebox{0.8}{%
    $\rmW^* = \left(\rmZ^\top\rmZ\right)^{-1}\rmZ^\top\rmY$
    }.
\end{equation}
The $i$-th row of $\rmZ$ is the normalized feature vector ${\rvz}_i$ corresponding to the $i$-th sample $\rvx_i$ in $\mathcal{D}$. Similarly, the $i$-th row of $\rmY$ is the one-hot target vector $\texttt{one\_hot}(y_i)$ corresponding to $\rvx_i$.

Obtaining $\rmZ$ (and $\rmY$) requires all clients to upload their features (and corresponding labels). 
However, in the context of federated learning, sharing features and labels will cause prohibitively expensive communication overheads and potential model inversion attack~\cite{dong2021deep,zhao2020makes}. 
To achieve efficient and privacy-enhanced calibration, we propose to compute $\rmW^*$ in a distributed manner. It is inspired by that matrix multiplication can be implemented by a sum of outer products~\cite{zhang2017box,liu2019multiple}.
\definecolor{LightGray}{gray}{0.90}
\definecolor{LightPGray}{gray}{0.73}
Take $\rmZ^\top\rmZ$ as an example. We can rewrite the $\rmZ^\top\rmZ$ as a sum of outer products between \colorbox{LightGray}{columns} of $\rmZ^\top$ and \colorbox{LightPGray}{\vphantom{l}rows} of $\rmZ$ such that
\begin{equation}
    \label{equ:outer_products}
    \rmZ^\top\rmZ = 
    \underbrace{%
    \tikzmarknode{n}{\colorbox{LightGray}{${\rvz}_1^{\top}$}}
    \otimes
    \tikzmarknode{n}{\colorbox{LightPGray}{\vphantom{${\rvz}_1^{\top}$}
    ${\rvz}_1$}} +\ \dotsi
    }_{\textbf{Client 1}}
    \ \dotsi\ 
    \underbrace{%
    \vphantom{a_{a_{a_{a_a}}}^b} 
    \dotsi\ 
    +\ 
    \dotsi}_{\textbf{Client k}}
    \ \dotsi\ 
    {\underbrace{%
    \vphantom{a^b}
    \dotsi + 
    \tikzmarknode{n}{\colorbox{LightGray}{
    \vphantom{${\rvz}_{1}^{\top}$}$\smash{{\rvz}_{\mathsmaller{\mathsmaller{|\mathcal{D}|}}}^{\top}}$}}
    \otimes
    \tikzmarknode{n}{\colorbox{LightPGray}{
    \vphantom{${\rvz}_{1}^{\top}$}$\smash{{\rvz}_{\mathsmaller{\mathsmaller{|\mathcal{D}|}}}}$}}
    }_{\textbf{Client K}}}~.
\end{equation}
As a result, to calculate $\rmZ^\top\rmZ$, each client can complete a fraction of sum over their local features (as shown in~\cref{equ:outer_products}) and upload the intermediate results to the sever to finish the computation, rather than upload private local features ${\rvz}_i$.
In addition, the computation of $\rmZ^\top\rmY$ in~\cref{equ:closed_form} can be calculated in the same way. See \cref{fig:teaser} (\textit{Right}) for a vivid illustration. 

\fakeparagraph{Algorithm.} We elaborate the FFC algorithm below:
\begin{enumerate}
    \item On clients: Each client receives the latest global feature extractor $f_\theta$ from the server and computes the $\rmV^k\in\mathbb{R}^{l\times l},\ \rmU^k\in\mathbb{R}^{l\times C}$ on its local data $\mathcal{D}^k$,
    \begin{equation}
        \label{equ:ss_v_u}
        \scalebox{0.9}{%
        $\rmV^k = \sum_{i=1}^{|\mathcal{D}^k|}\, {\rvz}_i^\top {\rvz}_i, \quad \rmU^k = \sum_{i=1}^{|\mathcal{D}^k|} {\rvz}_i^\top \texttt{one\_hot}(y_i)$
        }, 
    \end{equation}
    where ${\rvz}_i = f_\theta(\rvx_i)/\|f_\theta(\rvx_i)\|$ and $(\rvx_i, y_i) \sim \mathcal{D}^k$.
    
    \item On server: The server receives all $\{(\rmV^k, \rmU^k)\ |\ k\in[K]\}$ from clients and computes the closed-form weights optimum,
    \begin{equation}
        \label{equ:final_closed_form}
        \scalebox{0.85}{%
        $\rmW^* = \left(\sum_{k=1}^{K}\rmV^k\right)^{-1}\left(\sum_{k=1}^{K}\rmU^k\right)$
        }.
    \end{equation}
\end{enumerate}

\setlength{\tabcolsep}{3pt}
\begin{table}[t]
\centering
\caption{Comparison of communication and computation costs for the classifier over 10 clients for 100 rounds. We assume that communicated weights are in 32-bit. The number of FLOPs is computed by considering both clients (``(C)'') and the server (``(S)''). Since all approaches require forward of the classifier, we exclude it from calculation. ``Every'' (and ``Once'') means communicating at every round (and only once). }
\setlength\tabcolsep{5.18pt}
 \resizebox{\linewidth}{!}{
\begin{tabular}{@{\extracolsep{4pt}}c|lcc|cc|c@{}}
\toprule
    \multirow{2}{*}{\bftab Method}&
    \multicolumn{3}{c|}{\bftab Communication} & \multicolumn{2}{c|}{\bftab Computation} & 
    \multirow{2}{*}{\bftab Accuracy} \\             
    &
    \begin{tabular}[c]{@{}c@{}}\bftab Objects \end{tabular} &
    \begin{tabular}[c]{@{}c@{}} \bftab Frequency \end{tabular} &
    \begin{tabular}[c]{@{}c@{}} \bftab Size \scalebox{.7}{(MB)} \end{tabular} 
    &\bftab Operation &
    \begin{tabular}[c]{@{}c@{}} \bftab \scalebox{.9}{FLOPs (G)} \end{tabular}\\
\noalign{\smallskip}
\hline
\noalign{\smallskip}
\bftab FedAvg & \scalebox{.8}{$\rmW\in\mathbb{R}^{l\times C}$} & \scalebox{.9}{Every} & \begin{tabular}[c]{@{}c@{}} 102 \end{tabular} & \begin{tabular}[c]{@{}c@{}}  {\scriptsize Update \scalebox{.8}{$\rmW$} (S)}\\{\scriptsize Avg (S)} \end{tabular}  & \scalebox{.9}{$1.9\times 10^4$} & 68.78\\
\noalign{\smallskip}
\hline
\noalign{\smallskip}
\multirow{3}{*}{\shortstack[lb]{\ \\ \ \\ \bftab CCVR}} &\scalebox{.8}{$\rmW\in\mathbb{R}^{l\times C}$} & \scalebox{.9}{Every} & \multirow{3}{*}{\begin{tabular}[c]{@{}c@{}}\ 758\end{tabular} } & \multirow{3}{*}{\begin{tabular}[c]{@{}c@{}} {\scriptsize GMM fitting (C)}\\{\scriptsize Sampling\&Tukey (S)}\\{\scriptsize Fine-tuning (S)} \end{tabular} }  &   \multirow{3}{*}{\scalebox{.9}{$2.1\times 10^4$}} & \multirow{3}{*}{69.14}\\
\cline{2-3}
&\,\scalebox{.8}{$\mathbf{\mu}\in\mathbb{R}^{l\times C}$} & \multirow{2}{*}{\scalebox{.9}{Once}} & & &&\\
&\scalebox{.8}{$\mathbf{\Sigma}\in\mathbb{R}^{l\times l\times C}$} & & & &\\
\noalign{\smallskip}
\hline
\noalign{\smallskip}
\multirow{2}{*}{\begin{tabular}[c]{@{}c@{}} \bftab \method\\\bftab (Ours)  \end{tabular}} &\scalebox{.8}{$\rmU\in\mathbb{R}^{l\times C}$} &  \multirow{2}{*}{\scalebox{.9}{Once}} & \multirow{2}{*}{\begin{tabular}[c]{@{}c@{}} 7\end{tabular}} & \multirow{2}{*}{\begin{tabular}[c]{@{}c@{}} {\scriptsize \cref{equ:ss_v_u} (C)}\\{\scriptsize \cref{equ:final_closed_form} (S)} \end{tabular} } & \multirow{2}{*}{\scalebox{.9}{$1.7\times 10^2$}} & \multirow{2}{*}{71.85}\\
 &\scalebox{.8}{$\rmV\in\mathbb{R}^{l\times l}$}&&&&\\
\bottomrule
\end{tabular}
}
\vspace{-4mm}
\label{tab:ccvr_fedavg}
\end{table}
\setlength{\tabcolsep}{1.4pt}

\vspace{-2mm}
\fakeparagraph{Communication and Computation.} \methodd introduces much lighter communication and computation overhead to clients/server compared with state-of-the-art calibration methods like CCVR~\cite{luo2021no}.
For instance, computing~\cref{equ:ss_v_u} on one client requires $2l|\mathcal{D}^k|(l+C)$ FOLPs which is less than the total computation of the classifier's local training for one round. In addition, the communication amount of \methodd is $l(l+C)$ parameters.  
In~\cref{tab:ccvr_fedavg}, we compare the communication and computation amount for the classifier of \methodd against baselines (\eg, FedAvg~\cite{mcmahan2017communication} and CCVR~\cite{luo2021no}) over 100 rounds with MobileNetV2 and CIFAR-100. As depicted in~\cref{tab:ccvr_fedavg}, \method and \methodd are more efficient than both FedAvg and CCVR in terms of communication and computation. 
In~\cref{sec:exp:hardware}, we also provide latency comparison measured on a real embedded hardware.

\section{Experiments}
\label{sec:experiments}

\subsection{Experimental Setup}
\label{sec:setup}
\subsubsection{Baselines.} Our proposed methods (\ie, \method and \methodd) are compatible with and complementary to several existing federated learning algorithms like FedAvg~\cite{mcmahan2017communication}, FedProx~\cite{li2020federated}, FedNova~\cite{wang2020tackling}, FedOpt~\cite{reddi2020adaptive} and so on. We refer to those algorithms as ``base algorithms'' in the remaining of this paper.  
We test five widely used models including a seven-layer ConvNet~\cite{yoon2021fedmix,zhang2022multi} and other modern deep architectures like MobileNetV2~\cite{sandler2018mobilenetv2}, ResNet18~\cite{he2016deep}, VGG13~\cite{kuangliu_vgg,simonyan2014very}, SENet~\cite{hu2018squeeze}. For all models, we refer to the last fully-connected layer as the classifier and all the other layers as the feature extractor. 

\fakeparagraph{Benchmarks.} Following prior literature~\cite{li2021federated,li2021model,yoon2021federated}, we consider two representative and challenging image classification tasks for federated learning, CIFAR-100~\cite{Krizhevsky09learningmultiple} and TinyImageNet~\cite{Le2015TinyIV}. CIFAR-100 has 50,000 training samples from 100 classes. Prior method~\cite{luo2021no} obtains relatively small improvement against base algorithms on CIFAR-100 because the virtual features sampled from the class-wise Gaussian Mixture Model (GMM) are less separable when the number of class increases, thereby constraining its practicality for realistic applications. In this work, we show that our proposed methods are able to achieve superior performance for such many-class classification tasks. 
For example, we evaluate our methods on TinyImageNet which is larger than CIFAR-100 in terms of the input size, the number of samples, and the number of classes, as a more challenging dataset. Empirical results indicate that our proposed methods provide consistent improvements across multiple base algorithms, model architectures and datasets.  

Like previous studies~\cite{chaoyanghe2020fedml,kairouz2019advances,li2021model}, we partition the training set of CIFAR-100 and TinyImageNet to $K$ clients according to a Dirichlet distribution with a concentration parameter $\alpha$ to simulate the data distribution of federated learning. The default number of client is set to $K=10$. A smaller concentration parameter will result in higher \noniid degree of partitioning. For example, when $\alpha=0.1$, one client could have less than ten data samples in some classes. We consider three different \noniid degrees for CIFAR-100 to study how data heterogeneity degree impacts methods' performance. For fair comparison, we use exactly the same partitioning for all methods. The original test sets of CIFAR-100 and TinyImageNet are used to measure the resulted global model's testing accuracy. 

\fakeparagraph{Implementation.} We use the SGD optimizer with a momentum 0.9 and a weight decay $10^{-5}$ for all approaches. Since we change the loss function from cross entropy to mean square error and these two loss functions have different magnitude, we tune the learning rate for both baselines and our methods using grid search. We note that \method and \methodd do not introduce any extra hyper-parameter to base algorithms. In addition, we observe that our methods are more robust to various learning rates than base algorithms in~\cref{sec:exp:abla}. 
For baselines with extra hyper-parameters, we either use the recommended values from their papers~\cite{luo2021no,wang2020tackling} or carefully tune them~\cite{reddi2020adaptive,li2020federated}. To tune hyper-parameters, we use a $15\%$ of training data for validation. We train all approaches for 100 rounds and decay the learning rate every round using a cosine annealing schedule~\cite{loshchilov2016sgdr}.
We use $B=64$ local batch size and $E=10$ local epochs unless otherwise stated. 
In~\cref{sec:app:extra_res}, we further test our methods on different federated learning settings by varying local training epochs, number of clients, clients' participating rate, and learning rate scheduling strategies, similar trends are observed as shown in the following sections.

\setlength{\tabcolsep}{4pt}
\begin{table}[t]
\centering
\caption{Accuracy (\%) on CIFAR-100 and TinyImageNet with different degrees of \noniid. ``\plus '' means applying a considered method (CCVR, BABU and Ours) to a base FL algorithm (FedAvg, FedProx, FedNova and FedOpt). ``$\uparrow$'' (and ``$\downarrow$'') means accuracy improvement (and degradation) compared with the corresponding base algorithm. 
}
\setlength\tabcolsep{10pt}
 \resizebox{\linewidth}{!}{
\begin{tabular}{cccccc}

\toprule
\noalign{\smallskip}
    \multirow{1}{*}{\begin{tabular}[c]{@{}c@{}} \bftab Model \end{tabular}} &
    \multirow{1}{*}{\begin{tabular}[c]{@{}c@{}} \bftab Method \end{tabular}} &
    \begin{tabular}[c]{@{}c@{}} \bftab IID \end{tabular} &
    \begin{tabular}[c]{@{}c@{}} $\boldsymbol{\alpha=0.5}$ \end{tabular} &
    \begin{tabular}[c]{@{}c@{}} $\boldsymbol{\alpha=0.1}$ \end{tabular} & \scalebox{1.0}{\bftab TinyImageNet} 
    \\          
\noalign{\smallskip}
\hline
\noalign{\smallskip}
\multirow{4}{*}{\begin{tabular}[c]{@{}c@{}} \bftab MobileNetV2 \end{tabular}} & FedAvg & 71.86  & 68.78 & 63.90 & 29.95\\
& \plus\ {\scriptsize CCVR} & 72.09~($\uparrow$\,0.23)  & 69.14~($\uparrow$\,0.36) & 64.05~($\uparrow$\,0.15) & 31.41~($\uparrow$\,1.46)\\
& \plus\ {\scriptsize BABU} & 71.84~($\downarrow$\,0.02)  & 69.35~($\uparrow$\,0.57) & 64.91~($\uparrow$\,1.01)  & 28.38~($\downarrow$\,1.57)\\
& \plus\ \ Ours & {\bftab 73.56\,\,($\uparrow$\,1.72)} & {\bftab 71.85\,\,($\uparrow$\,3.07)} & {\bftab 66.52\,\,($\uparrow$\,2.62)} & {\bftab 34.72~($\uparrow$\,4.76)}\\
\noalign{\smallskip}
\hline
\noalign{\smallskip}
\multirow{4}{*}{\begin{tabular}[c]{@{}c@{}} \bftab ResNet \end{tabular}} & \scalebox{.9}{FedProx} & 70.19 & 67.50 & 65.63 & 30.55\\
& \plus\ {\scriptsize CCVR} & 71.31~($\uparrow$\,0.12) & 67.89~($\uparrow$\,0.39) & 66.09~($\uparrow$\,0.46) & 32.56~($\uparrow$\,2.01)\\
& \plus\ {\scriptsize BABU} & 71.66~($\uparrow$\,1.47) & 69.62~($\uparrow$\,2.12) & 67.90~($\uparrow$\,2.27) & 31.87~($\uparrow$\,0.32)\\
& \plus\ \ Ours & {\bftab 73.41~($\uparrow$\,3.22)} & {\bftab 72.20~($\uparrow$\,4.70)} & {\bftab 69.19~($\uparrow$\,3.56)}  & {\bftab 35.21~($\uparrow$\,4.66)}\\
\noalign{\smallskip}
\hline
\noalign{\smallskip}
\multirow{4}{*}{\begin{tabular}[c]{@{}c@{}} \bftab VGG13 \end{tabular}} & \scalebox{.9}{FedNova} & 62.12 & 60.49 & 57.20 & 39.63\\
& \plus\ {\scriptsize CCVR} & 62.53~($\uparrow$\,0.41) & 61.61~($\uparrow$\,1.12) & 58.13~($\uparrow$\,0.93) & 40.12~($\uparrow$\,0.49)\\
& \plus\ {\scriptsize BABU} & 62.03~($\downarrow$\,0.09) & 60.54~($\uparrow$\,0.05) & 58.95~($\uparrow$\,1.75) & 40.87~($\uparrow$\,1.24)\\
& \plus\ \ Ours             & {\bftab 65.50\,\,($\uparrow$\,3.38)} & {\bftab 65.12\,\,($\uparrow$\,4.63)} & {\bftab 62.54\,\,($\uparrow$\,5.34)} & {\bftab 45.21\,\,($\uparrow$\,5.58)}\\
\noalign{\smallskip}
\hline
\noalign{\smallskip}
\multirow{4}{*}{\begin{tabular}[c]{@{}c@{}} \bftab SENet \end{tabular}} & FedOpt & 61.89 & 59.60 & 57.46 & 24.29\\
& \plus\ {\scriptsize CCVR} & 61.97~($\uparrow$\,0.08) & 60.42~($\uparrow$\,0.82) & 57.93~($\uparrow$\,0.47) & 25.01~($\uparrow$\,0.72)\\
& \plus\ {\scriptsize BABU} & 62.27~($\uparrow$\,0.38) &  59.69~($\uparrow$\,0.09) & 56.75~($\downarrow$\,0.71) & 25.34~($\uparrow$\,1.05)\\
& \plus\ \ Ours & {\bftab 65.15\,\,($\uparrow$\,3.26)} & {\bftab 65.69\,\,($\uparrow$\,6.09)} & {\bftab 62.61\,\,($\uparrow$\,5.15)} & {\bftab 29.84\,\,($\uparrow$\,5.55)}\\
\bottomrule
\end{tabular}
 }
\vspace{-6mm}
\label{tab:main_res_tab}
\end{table}
\setlength{\tabcolsep}{1.4pt}

\subsection{Results}
\label{sec:main_res}

We present in~\cref{tab:main_res_tab} the test accuracy of various base algorithms before and after applying our methods (\ie, \method and \methodd) and two state-of-the-art baselines (\ie, CCVR~\cite{luo2021no} and BABU~\cite{oh2021fedbabu}). 
Our proposed methods improve these base algorithms consistently across model architectures and datasets. 

CCVR estimates Gaussian Mixture Model (GMM) for features on the class granularity on clients and samples virtual features from the GMM to fine-tune the classifier on the server. However, when the number of classes is relatively large (\eg, 100 for CIFAR-100 and 200 for TinyImageNet), class-wise GMMs are not sufficiently separable to facilitate the fine-tuning. As a result, the improvement brought by CCVR is relatively small~\cite{luo2021no}. 

BABU~\cite{oh2021fedbabu} keeps the classifier fixed after random initialization during federated training and then fine-tunes on each client's local dataset individually for personalization. Two evaluation metrics are considered in BABU: (\romannumeral1) initial accuracy which is calculated with the single global model on the global test set and (\romannumeral2) personalized accuracies measured with personalized models on local test sets over clients. As mentioned previously, we focus on the former (\ie, initial accuracy) rather than personalized federated learning, while an analysis on how our method helps personalized federated learning is provided in~\cref{sec:app:pfl}.

An interesting finding is that our methods tend to bring more accuracy gain when the \noniid degree is higher. This confirms our observation that a higher \noniid degree leads to more severe issues on classifier disparity and inconsistent local learning targets. With our methods, the performance gap between \iid and \noniid data is reduced. For instance, the performance gap between ``IID'' and ``$\alpha=0.1$'' is $4.92\%$ for base algorithm FedNova, while this gap is reduced to $2.96\%$ after applying the proposed approaches.

\setlength{\tabcolsep}{8pt}
\begin{table}[t]
\centering
\caption{Quantitative ablation study of Hyperspherical Federated Learning (\method). We investigate the effectiveness of each design component by applying them individually using FedAvg as the base algorithm on MobileNetV2 and CIFAR-100 ($\alpha=0.5$). ``Fix (R)'' (``Fix (OU)'') means fixing the classifier with random (orthogonal and unit-norm) initialization. ``Norm'' represents normalizing features.}
\vspace{-2mm}
\label{tab:our_breakdown}
 \resizebox{\linewidth}{!}{
\begin{tabular}{cccccc}
\toprule
    & \begin{tabular}[c]{@{}c@{}} {\bftab FedAvg} \end{tabular} &
    \begin{tabular}[c]{@{}c@{}} {\bftab\plus\ Fix (R)} \end{tabular} &
    \begin{tabular}[c]{@{}c@{}} {\bftab\plus\ Fix (OU)} \end{tabular} &
    \begin{tabular}[c]{@{}c@{}} {\bftab\plus\ Norm} \end{tabular} &
    \begin{tabular}[c]{@{}c@{}} {\bftab\plus\ Fix (OU) \plus Norm} \end{tabular}
    \\ 
\noalign{\smallskip}
\hline
\noalign{\smallskip}
{\bftab CE} & 68.78 & 69.35 ($\uparrow$\,0.57) & 69.65 ($\uparrow$\,0.87) & 69.76 ($\uparrow$\,0.98) & 70.87 ($\uparrow$\,2.09)\\
\noalign{\smallskip}
{\bftab MSE} & 66.42 ($\downarrow$\,2.36) & 67.05 ($\downarrow$\,1.73) & 67.21 ($\downarrow$\,1.57) & Diverging & {\bftab 71.85 ($\uparrow$\,3.07)}\\
\bottomrule
\end{tabular}
}
\vspace{-4mm}
\end{table}

\subsection{Ablation Studies}
\label{sec:exp:abla}

Besides overall effectiveness, we perform several ablation experiments which help understand the significance of each component in the proposed method.

\fakeparagraph{The Importance of Ortho-normalization.}
In~\cref{tab:our_breakdown}, we first compare different initializations of the fixed classifier. For the orthogonal and unit-norm initialization~(\scalebox{.8}{``\plus\ Fix (OU)''}), we generate orthogonal weight matrix via the classic Gram-Schmidt process~\cite{torch_linalg_qr,pursell1991gram}. Other generation methods~\cite{tammes1930origin,mettes2019hyperspherical} are also considered in the appendix and no significant differences are observed. For the random initialization~(\scalebox{.8}{``\plus\ Fix (R)''}), we instantiate it with He Initialization~\cite{he2015delving} which is the default initialization method in widely used packages such as PyTorch~\cite{paszke2019pytorch}. 
Random initialization achieves a comparable but slightly lower accuracy gain because two random vectors tend to be more orthogonal when their dimensionality increases~\cite{lezama2018ole,oh2021fedbabu}, while orthogonal initialization directly ensures that. 

In addition, \method also normalizes features before feeding them to the classifier (denoted \scalebox{.9}{``Norm''} in~\cref{tab:our_breakdown}). After normalization, features are in the same unit hypersphere as the row vectors of classifier's weight and the feature extractor can focus on learning features' directions with the guidance of the fixed classifier. Applying feature normalization for the \scalebox{.85}{``Fix (OU)''} completes the construction of hyperspherical representation and leads to about 1.22 accuracy gain.
More importantly, we show that \scalebox{.85}{``Fix (OU)''} and feature normalization work better with MSE than CE in the following discussion.

\fakeparagraph{The Superiority of Mean-Square-Error Loss.}
We evaluate both CE and MSE loss functions in~\cref{tab:our_breakdown} to validate our choice of MSE loss. 
We confirm that replacing CE with MSE improves the accuracy by a considerable margin. The reasons are stated in~\cref{sec:mse_loss} that MSE avoids the scaling issue 
and fully exploits the benefit of \scalebox{.9}{``Fix (OU)\,\plus\,Norm''}.
For \scalebox{.9}{``CE\,\plus\,Fix (OU)\,\plus\,Norm''}, we find that it is quite sensitive to the scaling hyper-parameter (\ie, temperature). Although we carefully tune the scaling factor $\tau$ and report the best result in~\cref{tab:our_breakdown}, it is difficult and expensive to find the optimal $\tau$ in practice.  

\begin{figure}[t]
\begin{minipage}[b]{0.38\textwidth}
\centering
\includegraphics[width=\textwidth]{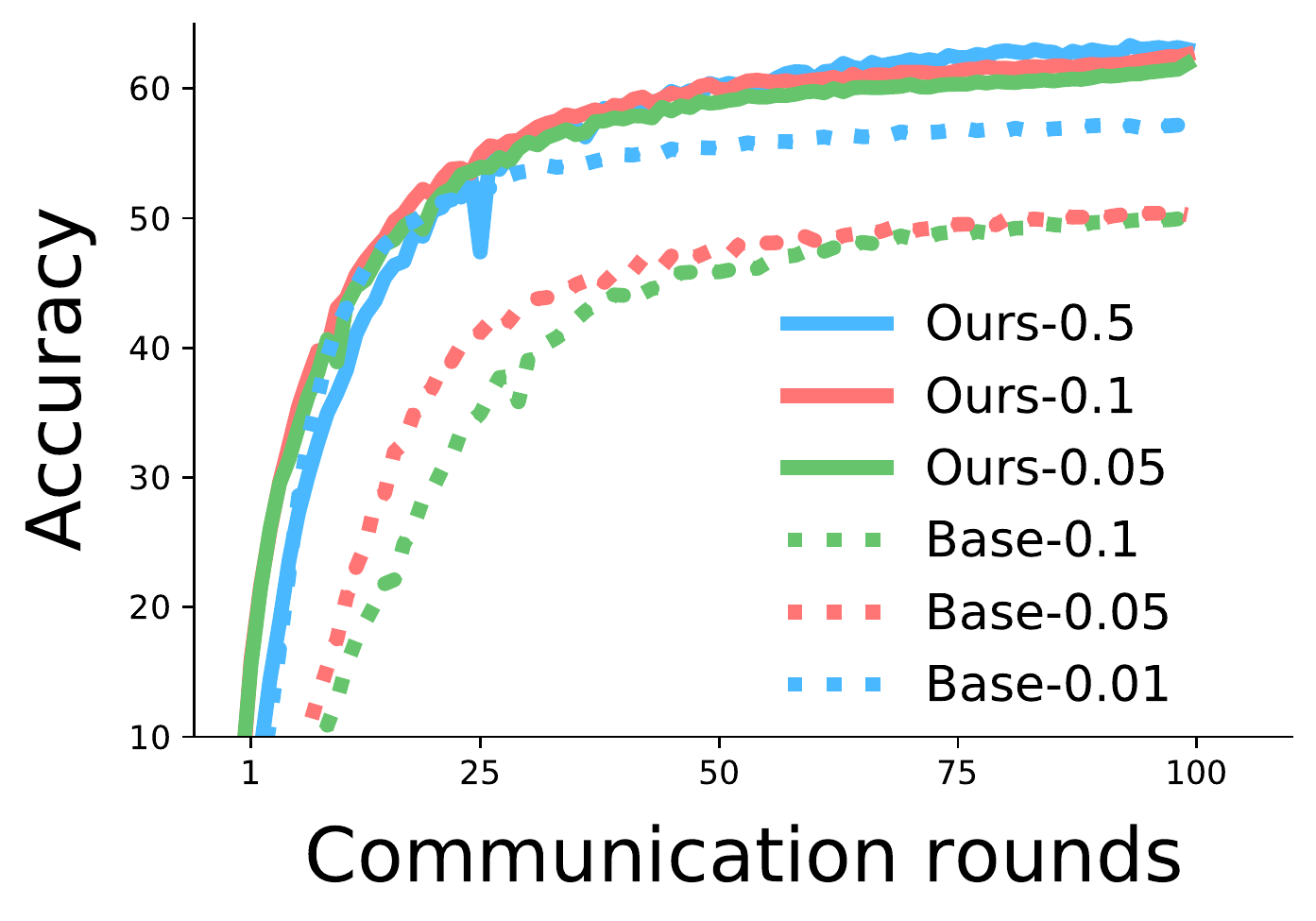}
\vspace{-6mm}
\captionof{figure}{The impact of different learning rates for ``FedNova'' and ``FedNova\,\plus\,\method''. After applying \method, training becomes more robust to different learning rates.}
\label{fig:lr_cruves}
\end{minipage}\hfill
\begin{minipage}[b]{0.57\textwidth}
\centering
\setlength{\belowcaptionskip}{10pt}
\captionof{table}{Ablation study for \methodd. Both CCVR and \methodd methods exhibit performance gain on \noniid data. For sanity check, we collect local train sets to fine-tune the classifier. \methodd is able to achieve a larger accuracy improvement than CCVR with significantly less communication and computation overheads (\cref{tab:ccvr_fedavg}).}
\label{tab:cali}
\resizebox{\textwidth}{!}{%
\begin{tabular}[b]{ccc}
\toprule
  \begin{tabular}[c]{@{}c@{}} {\bftab Calibration} \end{tabular} &
    \begin{tabular}[c]{@{}c@{}} {\bftab IID} \end{tabular} &
    \begin{tabular}[c]{@{}c@{}} {\bftab $\boldsymbol{\alpha=0.1}$} \end{tabular} \\

\noalign{\smallskip}
\hline
\noalign{\smallskip}
{\bftab W/o} & {65.07}  & 61.66\\
\noalign{\smallskip}
\ {\bftab\scalebox{.9}{CCVR}} & 65.03~($\downarrow$\,0.04) & 62.09~($\uparrow$\,0.43)\\
\noalign{\smallskip}
\ {\bftab\scalebox{.9}{\methodd}} & {\bftab 65.15~($\uparrow$\,0.08)} & {\bftab 62.61~($\uparrow$\,0.95)}\\
\noalign{\smallskip}
\hline
\noalign{\smallskip}
{\bftab\scalebox{.9}{Sanity check}} & {65.17~($\uparrow$\,0.10)} & {62.64~($\uparrow$\,0.98)} \\
\bottomrule
\end{tabular}
}
\end{minipage}

\vspace{-4mm}
\end{figure}

\fakeparagraph{The Robustness of \method Training.}
Compared with base algorithms, \method does not introduce any extra hyper-parameters. Since the CE and MSE losses have different magnitudes, we tune their learning rates respectively from a set of candidate learning rates.  Interestingly, we observe that \method is more robust than the corresponding base algorithm. In~\cref{fig:lr_cruves}, we test three different learning rates for \scalebox{0.9}{``FedNova''} (as the base algorithm) and \scalebox{0.9}{``FedNova\,\plus\,\method''} with VGG13 on CIFAR-100~($\alpha=0.1$). It is observed that \scalebox{0.9}{``FedNova\,\plus\,\method''} is less sensitive to different learning rates. 

\setlength{\tabcolsep}{9pt}

\fakeparagraph{How beneficial is \methodd?}
In this set of experiments, we investigate how the closed-form classifier calibration (\ie, \methodd) improves test accuracy. We apply CCVR and \methodd individually for the classifier of SENet on CIFAR-100 after federated learning with ``FedOpt\,\plus\,\method''. As a sanity check, we collect all local train sets to fine-tune the classifier only. To ensure the sanity check truly reveals the upper bound of classifier calibration, we experiment different loss functions (\ie, CE and MSE) and learning rates for the sanity fine-tuning and report the best results we get.
It can be seen from \cref{tab:cali} that both CCVR and \methodd achieve performance gains on \noniid data while \methodd is able to improve accuracy more. CCVR estimates a GMM distribution for each class's features and sample virtual features for model fine-tuning on the server. However, such a class-wise method has relatively high communication and computation complexities (which scale linearly with number of classes). Moreover, GMMs of different classes could be less separable when the number of classes increases, thereby further limiting its effectiveness. In contrast, \methodd, with provable formulations, is agnostic to number of classes and more suitable for realistic many-class federated learning tasks~\cite{hsu2020federated,yang2019quasi}. It is expected that \methodd obtains comparable results as the sanity check because solving a linear classifier with either closed-form equations or SGD will converge to similar optimums~\cite{boyd2004convex,saad1998online}.

\subsection{Efficient Communication and Computation}
\label{sec:exp:hardware}

\begin{figure}[t]
    \centering
    \includegraphics[width=\linewidth]{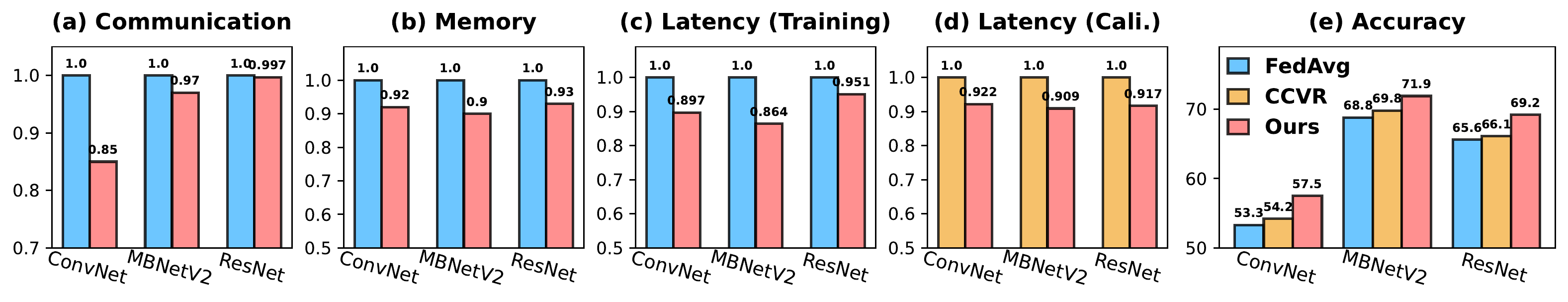}
    \vspace{-7mm}
    \caption{Efficiency comparison measured on a neural network accelerator~\cite{fpga,zhang2021fast} with three models and CIFAR-100. (a) The normalized total communication amount. (b) The normalized peak memory consumption during local training. (c) The normalized latency for one-round local training. (d) The normalized latency for classifier calibration. (e) The test accuracy.}
    \label{fig:hardware}
    \vspace{-4mm}
\end{figure}

Besides accuracy gain, Hyperspherical Federated Learning also brings communication and computation savings depending on the size of classifier. 

In~\cref{tab:ccvr_fedavg}, we compare the communication and computation costs related to the classifier for FedAvg, CCVR and our methods on MobileNetV2 and CIFAR-100. As seen in~\cref{tab:ccvr_fedavg}, \method eliminates the need of updating and communicating (neither uploading nor downloading), resulting over two orders of magnitude communication and computation savings compared to FedAvg and CCVR. 
\Cref{fig:hardware} depicts the relative computational savings of different approaches during the federated training process, measured on a DNN training accelerator built on Xilinx
VC707 FPGA evaluation board~\cite{fpga,zhang2021fast}. Detailed settings about the DNN training accelerator are described in the appendix. \method enables us to skip gradient computing for the classifier and thus to release some intermediate tensors used by gradient computing earlier. Overall, \method achieves up to 10.5\% and 13.6\% savings on memory consumption and processing latency compared with FedAvg, respectively. \method achieves a greater saving of training latency on MobileNetV2 than ConvNet and ResNet. This is because the computation workload associated with the convolutaional layers are smaller in MobileNetV2 due to the usage of the depthwise separable operations~\cite{sandler2018mobilenetv2}, leading to a greater relative savings when the classifier is skipped during the local training. We also measure the latency of classifier calibration using CCVR and our Fast Federated Calibration in \cref{fig:hardware} (d). Since calibration-related computation happens on both clients and the server, the reported latency consists of both the average latency on one client and the on-server latency. Our closed-form calibration saves up to 9.1\% latency against CCVR and this efficiency improvement will be more pronounced when number of classes increases as analysed above.

\section{Conclusions}
We presented the Hyperspherical Federated Learning (\method) framework to address the \noniid issue. The proposed method focuses on the pre-learning phase and complementary to existing federated learning methods. The hyperspherical representation is learned using an orthonormal basis of the weights of the frozen classifiers and the classifiers are calibrated post training. We show that a mean squared loss is more suitable to hyperspherical representation as opposed to cross-entropy due to the scaling issues. A Fast Federated Calibration (FFC) approach is proposed based on the mean squared loss. Extensive experiments indicate that \method improves multiple existing federated learning algorithms by a considerable margin.

\clearpage
\bibliographystyle{splncs04}
\bibliography{egbib}

\clearpage
\appendix

\section{Extra Federated Learning Results}
\label{sec:app:extra_res}
In this section, we present additional results under various system and training settings to further evaluate the robustness of our approach.  

\subsection{Different Local Training Epochs}

Tuning the number of local epochs affects accuracy-communication trade-off for most federated learning algorithms. 
Prior studies~\cite{karimireddy2020scaffold,mcmahan2017communication,reddi2020adaptive} attempt to reduce the number of local training epochs to mitigate the disparity of local models. The default number of local training epochs is set as $E=10$ in the main manuscript. We further test different numbers of local training epochs $E=\{1,5\}$ in~\cref{tab:app:local_ep}. When $E$ is set as 1, we increase the total number of rounds to 500 to ensure convergence. 
As is seen from~\cref{tab:app:local_ep}, our method consistently improves the base algorithms with various numbers of local training epochs. 

\begin{table}[h]
\centering
\caption{Accuracy (\%) with different number of local training epochs.}
\setlength\tabcolsep{16pt}
\begin{adjustbox}{max width=\textwidth}
\begin{tabular}{cccc}

\toprule
\noalign{\smallskip}
    \multirow{1}{*}{\begin{tabular}[c]{@{}c@{}} \bftab Model \end{tabular}} &
    \multirow{1}{*}{\begin{tabular}[c]{@{}c@{}} \bftab Method \end{tabular}} &
    \begin{tabular}[c]{@{}c@{}} \bftab $\boldsymbol{E=1}$ \end{tabular} &
    \begin{tabular}[c]{@{}c@{}} $\boldsymbol{E=5}$ \end{tabular}

    \\          
\noalign{\smallskip}
\hline
\noalign{\smallskip}
\multirow{4}{*}{\makecell{{MobileNetV2}\\ (${\alpha=0.5}$)}} & FedAvg & 70.51 & 68.40 \\
& \plus\ {\scriptsize CCVR} & 71.33\,\,($\uparrow$\,0.82) & 68.93\,\,($\uparrow$\,0.53)\\
& \plus\ {\scriptsize BABU} & 71.46\,\,($\uparrow$\,0.95) & 69.10\,\,($\uparrow$\,0.70)\\
& \plus\ \ Ours & {\bftab 73.26\,\,($\uparrow$\,2.75)} & {\bftab 72.02\,\,($\uparrow$\,3.62)}\\
\noalign{\smallskip}
\hline
\noalign{\smallskip}
\multirow{4}{*}{\makecell{{VGG13}\\ (${\alpha=0.1}$)}} & \scalebox{0.9}{FedNova} & 60.53 & 57.46 \\
& \plus\ {\scriptsize CCVR} & 61.31\,\,($\uparrow$\,0.78) & 57.77\,\,($\uparrow$\,0.31)\\
& \plus\ {\scriptsize BABU} & 62.75\,\,($\uparrow$\,2.22) & 58.50\,\,($\uparrow$\,1.04) \\
& \plus\ \ Ours & {\bftab 64.75\,\,($\uparrow$\,4.22)} & {\bftab 62.06\,\,($\uparrow$\,4.60)} \\
\bottomrule
\end{tabular}
\end{adjustbox}
\vspace{-6mm}
\label{tab:app:local_ep}
\end{table}
\subsection{Different Client Numbers}
The default number of clients $K$ is set as 10 following prior work~\cite{li2021model,luo2021no}. We further test the performance of our methods when the system contains more clients. In~\cref{tab:app:more_clients}, we partition CIFAR-100 training set to $K=100$ clients according to a Dirichlet distribution with a concentration parameter $\alpha=0.5$. During federated training, the central server randomly selects 10\% clients to participate each round~\cite{karimireddy2020scaffold,li2020federated,oh2021fedbabu,reddi2020adaptive}. For each method, we set the number of rounds as 500. According to results in~\cref{tab:app:more_clients}, 

\begin{table}[h]
\centering
\caption{Accuracy (\%) with $K=100$ clients.}
\setlength\tabcolsep{10pt}
\begin{adjustbox}{max width=\textwidth}
\begin{tabular}{ccc||ccc}

\toprule
\noalign{\smallskip}
    \multirow{1}{*}{\begin{tabular}[c]{@{}c@{}} \bftab Model \end{tabular}} &
    \multirow{1}{*}{\begin{tabular}[c]{@{}c@{}} \bftab Method \end{tabular}} &
    \begin{tabular}[c]{@{}c@{}} \bftab Accuracy \end{tabular} &
    \begin{tabular}[c]{@{}c@{}} \bftab Model \end{tabular} &
    \begin{tabular}[c]{@{}c@{}} \bftab Method \end{tabular} & 
    \begin{tabular}[c]{@{}c@{}} \bftab Accuracy \end{tabular}
    \\          
\noalign{\smallskip}
\hline
\noalign{\smallskip}
\multirow{4}{*}{\scalebox{0.8}{MobileNetV2}} & FedAvg & 68.26 &  \multirow{4}{*}{\scalebox{0.8}{VGG13}} & \scalebox{0.9}{FedNova} & 49.04\\
& \plus\ {\scriptsize CCVR} & 69.20\,\,($\uparrow$\,0.94) & & \plus\ {\scriptsize CCVR} & 50.45\,\,($\uparrow$\,1.41)\\
& \plus\ {\scriptsize BABU} & 69.14\,\,($\uparrow$\,0.88) & & \plus\ {\scriptsize BABU} & 51.27\,\,($\uparrow$\,2.23)\\
& \plus\ \ Ours & {\bftab 71.38\,\,($\uparrow$\,5.12)} & & \plus\ \ Ours & {\bftab 55.23\,\,($\uparrow$\,6.19)}\\

\bottomrule
\end{tabular}
\end{adjustbox}
\vspace{-6mm}
\label{tab:app:more_clients}
\end{table}

\subsection{Different Learning Rate Scheduling Strategies}
Besides adjusting the learning rate at each round according to a cosine annealing schedule\footnote{See \texttt{torch.optim.lr\_scheduler.CosineAnnealingLR}}, we further experiment another widely used learning rate scheduling strategy (\eg, multi-step scheduling\footnote{See \texttt{torch.optim.lr\_scheduler.MultiStepLR}}) to verify the robustness of our approaches. Specifically, we decay the learning rate by $0.1$ every $40$ epochs. Empirical results in~\cref{tab:app:multi_step} indicate that our methods are able to bring state-of-the-art accuracy gain with different learning rate scheduling strategies.  
\begin{table}[h]
\centering
\caption{Accuracy (\%) with different learning rate (LR) schedulings.}
\setlength\tabcolsep{10pt}
\begin{adjustbox}{max width=\textwidth}
\begin{tabular}{ccc||ccc}

\toprule
\noalign{\smallskip}
    \multirow{1}{*}{\begin{tabular}[c]{@{}c@{}} \bftab LR  \end{tabular}} &
    \multirow{1}{*}{\begin{tabular}[c]{@{}c@{}} \bftab method \end{tabular}} &
    \begin{tabular}[c]{@{}c@{}} \bftab Accuracy \end{tabular} &
    \begin{tabular}[c]{@{}c@{}} \bftab LR \end{tabular} &
    \multirow{1}{*}{\begin{tabular}[c]{@{}c@{}} \bftab Method \end{tabular}} &
    \begin{tabular}[c]{@{}c@{}} \bftab Accuracy \end{tabular}
    \\          
\noalign{\smallskip}
\hline
\noalign{\smallskip}
\multirow{4}{*}{\scalebox{1}{Cosine}} & FedAvg & 68.78 &  \multirow{4}{*}{\scalebox{1}{Multi-step}} & \scalebox{1}{FedAvg} & 68.60\\
& \plus\ {\scriptsize CCVR} & 69.14\,\,($\uparrow$\,0.36) & & \plus\ {\scriptsize CCVR} & 69.05\,\,($\uparrow$\,0.45)\\
& \plus\ {\scriptsize BABU} & 69.35\,\,($\uparrow$\,0.57) & & \plus\ {\scriptsize BABU} & 69.43\,\,($\uparrow$\,0.83)\\
& \plus\ \ Ours & {\bftab 71.85\,\,($\uparrow$\,3.07)} & & \plus\ \ Ours & {\bftab 71.06\,\,($\uparrow$\,2.46)}\\
\bottomrule
\end{tabular}
\end{adjustbox}
\vspace{-6mm}
\label{tab:app:multi_step}
\end{table}

\section{Personalization Performance Comparison}
\label{sec:app:pfl}
We investigate the performance gain brought by \method for personalized federated learning. Following the setup in~\cite{jiang2019improving,oh2021fedbabu,wang2019federated}, we first combine the training and testing sets of CIFAR-100 to a single dataset which contains 60,000 samples in total. Then, the combined dataset is partitioned into 10 clients according to a Dirichlet distribution with a concentration parameter $\alpha$. On each client, we use $15\%$ its local data as local testing set and the other as local training set. Each personalized local model is evaluated on its corresponding local testing set. The overall performance of personalized federated learning is evaluated by calculating the mean and standard deviation of local testing accuracies across all clients~\cite{jiang2019improving,oh2021fedbabu,wang2019federated}. 

We consider four recent personalized federated learning baselines for comparison in~\cref{tab:pfl_tab}. For instance, LG-FedAvg~\cite{liang2020think} jointly updates feature extractors and classifiers during local training and only aggregates classifiers on the server in order to learn compact local representations. In FedRep~\cite{collins2021exploiting}, local feature extractors and classifiers are updated sequentially and the servers aggregates updated feature extractors at each round. The state-of-the-art pFL method, BABU~\cite{oh2021fedbabu}, is also included in the comparison. 

For our methods, we use \method during federated training. Since it is not necessary to get the optimal global classifier for pFL, we skip the \methodd for the global classifier. Instead, we keep the learnt global feature extractor fixed and conduct personalization fine-tuning for the local classifier on each client. We consider two manners for the personalization fine-tunings. The first manner (duded `Ours (SGD)' in~\cref{tab:pfl_tab}) is to optimize the classifier on the local training set with SGD optimizer like prior arts~\cite{jiang2019improving,oh2021fedbabu,wang2019federated}. The second manner (duded `Ours (\methodd)' in~\cref{tab:pfl_tab}) is to adapt our \methodd to compute the closed-form optimal classifier on the local training set (according to~\cref{equ:closed_form}). 

Empirical results in~\cref{tab:pfl_tab} indicate that our proposed methods are able to improve pFL as well with hyperspherical features which are better aligned and less biased. 
\setlength{\tabcolsep}{8pt}
\begin{table}[h]
\centering
\caption{Accuracy (\%) comparison of different personalized federated learning (pFL) methods. The numbers `A\,$\pm$\,B' are the `mean\,$\pm$\,standard deviation' of personalized accuracies across clients.}
\resizebox{\linewidth}{!}{
\begin{tabular}{ccccccc}
\toprule
\noalign{\smallskip}
    \multirow{1}{*}{\begin{tabular}[c]{@{}c@{}} {\bftab Data} \end{tabular}} &
    \multirow{1}{*}{\begin{tabular}[c]{@{}c@{}} {\bftab FedAvg} \end{tabular}} &
    \begin{tabular}[c]{@{}c@{}} {\bftab LG-FedAvg} \end{tabular} &
    \begin{tabular}[c]{@{}c@{}} {\bftab FedRep} \end{tabular} &
    \begin{tabular}[c]{@{}c@{}} {\bftab BABU} \end{tabular} & \scalebox{1.0}{\bftab Ours (\methodd)} & \scalebox{1.0}{\bftab Ours (SGD)} 
    \\          
\noalign{\smallskip}
\hline
\noalign{\smallskip}
$\boldsymbol{\alpha=0.5}$ & $69.95\pm3.96$ & $71.67\pm4.33$ & $62.39\pm3.91$ & $72.34\pm3.84$ & $75.63\pm3.13$ & {\bftab 75.71$\pm$3.32} \\
\noalign{\smallskip}
$\boldsymbol{\alpha=0.1}$ & $80.16\pm3.77$ & $81.95\pm3.69$ & $73.63\pm3.74$ & $82.12\pm3.63$ & $84.06\pm3.10$ & {\bftab 84.12$\pm$3.29} \\
\bottomrule
\end{tabular}
}
\label{tab:pfl_tab}
\end{table}

\section{Extra Calibration Results}

\subsection{Adapting An $\ell_2$ Penalty in the Closed-Form Solution}
In \cref{sec:mse_loss}, we formulate the calibrating the classifier $\rmW$ as a least square problem in \cref{equ:loss_for_w}. Theoretically, an $\ell_2$ penalty of $\rmW$ can be added and the objective of calibrating the classifier is,
\begin{equation}
\label{equ:loss_for_w_l2}
    \scalebox{0.85}{%
    $\argmin_\rmW\ \mathbb{E}_{(\rvx, y)\sim\mathcal{D}}\left[\mathcal{L}_\textrm{{\tiny MSE}}\left(\rmW\rvz,\ y\right)\right]+\lambda\|\rmW\|^2_2$
    }~,
\end{equation}
where $\lambda$ is a hyper-parameter used to control the penalty intensity. As a result, the closed-form weights optimum (\ie, \cref{equ:final_closed_form}) becomes,
\begin{equation}
    \label{equ:final_closed_form_l2}
    \scalebox{0.85}{%
    $\rmW^* = \left(\sum_{k=1}^{K}\rmV^k+\lambda\rmI\right)^{-1}\left(\sum_{k=1}^{K}\rmU^k\right)$
    },
\end{equation}
where $\rmI\in\mathbb{R}^{l\times l}$ is the identity matrix.
In practise, we test different values for $\lambda\in\{0,\ 10^{-3},\ 10^{-2},\ 10^{-1},\ 10^{0},\ 10^{1}\}$ and find that the difference among resulted accuracies is less than 0.27\% ($10^{-1}$ results in the better accuracy 71.96\%). Therefore, we keep $\lambda=0$ in our main experiments. 

\subsection{Calibrate Weights One or Multiple Times?}
Theoretically, we can conduct the Fast Federated Calibration~(\methodd) multiple times during the federated training. In practise, we attempt to calibrate the classifier every 10 rounds and get a final accuracy 71.88\% which is quite close to the accuracy of one-time calibration (71.85\%). This empirical result suggests that the pre-defined orthogonal hyperspherical $\rmW$ serves as a high-quality feature learning target (as discussed in \cref{sec:hyper_rep}) against a calibrated one.
Moreover, conducting calibration multiple times will introduce extra communication and computation overheads. In this regard, we conduct \methodd once after federated training in our main experiments. 

\subsection{Applying \methodd with Features Trained with CE}
The derivation of Fast Federated Calibration (\methodd) relies on the mean square error (MSE) loss. However, in this section, we show that \methodd can be used upon features trained by other losses (\eg, cross entropy loss) because the training of the feature extractor and the calibration of the classifier are decoupled. In addition, both CE and MSE have similar optimization goal, \ie, encouraging the feature extractor to make features of the $i$-th class close to $\rvw_i$.  
For instance, we train the feature extractor with cross entropy loss (CE) and then calibrate the classifier with our proposed \methodd in~\cref{tab:app:cali_ce}. Experiential results verify that \methodd is able to improve the classifier on features trained with CE.

\setlength{\tabcolsep}{8pt}
\begin{table}[h]
\centering
\caption{Applying \methodd for the classifier on features trained with CE.}
\label{tab:app:cali_ce}
\begin{tabular}{cccc}
\toprule
    & \begin{tabular}[c]{@{}c@{}} {\bftab FedAvg} \end{tabular} &
    \begin{tabular}[c]{@{}c@{}} {\bftab\plus\ Fix (R)} \end{tabular} &
    \begin{tabular}[c]{@{}c@{}} {\bftab\plus\ Fix (R) \plus\ \methodd} \end{tabular}
    \\ 
\noalign{\smallskip}
\hline
\noalign{\smallskip}
{\bftab CE} & 63.90 & 64.91 ($\uparrow$\,1.01) & 65.36 ($\uparrow$\,1.46) \\
\bottomrule
\end{tabular}
\end{table}

\section{Different Ways to Generate Orthogonal Classifier Initialization}

We experiment two representative methods to generate the row-orthogonal weight matrix for classifier's weight initialization. 
\begin{itemize}
    \item QR-decomposition method. In linear algebra, a QR decomposition is to decompose a random matrix $\rmA$ into a product $\rmA = \rmQ\rmR$ of an orthogonal matrix $\rmQ$ and an upper triangular matrix $\rmR$~\cite{torch_linalg_qr,pursell1991gram,trefethen1997numerical}, in which $\rmQ$ is the matrix of our interest. 
    \item Tammes method. To distribute $C$ two-dimensional vectors on an unit circle as uniformly as possible, one can randomly place the first vector and then put the next vector by shifting the previous vector for an angle of $\frac{2\pi}{C}$. However, when the dimension is larger than two, no such optimal separation algorithm exists, which is known as the Tammes problem~\cite{tammes1930origin}. To maximize the separation for any vector dimension, Mettes~\etal~\cite{mettes2019hyperspherical} optimize an objective which encourages large cosine similarity of any pair of vectors with gradient decent. Following~\cite{mettes2019hyperspherical}, we learn the vectors using SGD optimizer with $0.1$ learning rate and 0.9 momentum for $10^4$ steps.  
\end{itemize}

\setlength{\tabcolsep}{6pt}
\begin{table}[h]
\centering
\caption{Accuracy (\%) of different initialization methods for the classifier.}
\begin{tabular}{cccccc}
\toprule
\multirow{3}{*}{\begin{tabular}[c]{@{}c@{}}\vspace{1mm}{\bftab \scriptsize FedAvg+\method} \\ {\bftab \scriptsize (MobileNetv2}\\{\bftab \scriptsize on CIFAR-100)}\end{tabular} } & {\bftab Init.} & {\bftab Rep. dim. ($l$)} & {\bftab \#Classes ($C$)} & {\bftab Time} & {\bftab Accuracy} \\
\noalign{\smallskip}
\cline{2-6}
\noalign{\smallskip}
& QR & 1280  & 100 & 0.02~s & 71.85\\ 
& Tammes & 1280 & 100 & 13.1~s & 71.36\\
\bottomrule
\end{tabular}
\label{tab:orth_init_comp}
\end{table}
\setlength{\tabcolsep}{1.4pt}

\cref{tab:orth_init_comp} shows the comparison of above two kinds of initialization methods for the orthogonal classifier weight matrix. QR-decomposition initialization achieves slightly better accuracy than the Tammes initialization. We also provide the wall time of the two methods which is measured on the machine with one NVIDIA GeForce GTX 1080 Ti GPU.

In all the other experiments of this work, QR method is used for \method due to its efficiency and effectiveness.

\section{Details of Hardware Experiments}
We evaluate the hardware performance of different on an embedded DNN training accelerator~\cite{zhang2021fast} based on the Xilinx VC707 FPGA evaluation board~\cite{fpga}.
A 32 by 64 systolic array is used to perform the tensor operations during the forward and backpropagation. Each systolic cell consists of a Multiply-Accumulate (MAC) unit which can perform a floating-point multiplication and addition within a clock cycle, and a special unit is implemented to perform the operations for the rest layers (\eg, group/batch normalization, ReLU). The hardware system runs at 100 MHz.

\section{Details of Implementation}

\subsection{Model Architectures}
We provide the detailed information of ConvNet, MobileNetV2, ResNet18, VGG13 and SENet18 in~\cref{tab:model:convnet,tab:model:mbv2,tab:model:resnet,tab:model:vgg,tab:model:senet}. For the `NormLayer' after each convolution layer, two kinds of normalization layers are experimented. For experiments with MobileNetV2 and CIFAR-100, we instance the normalization layer as batch normalization. For other experiments, we use group normalization following prior arts~\cite{karimireddy2020scaffold,li2020federated,luo2021no,shoham2019overcoming,wang2020tackling,yoon2021federated,yuan2021federated}.

\subsection{Hyper-parameters}
\method and \methodd do not introduce any extra hyper-parameter to base federated learning algorithms.
Since we change the loss function from
cross entropy to mean square error and these two loss functions have different
magnitude, we tune the learning rate for both baselines and our methods using grid search from the limited candidate set $\{0.005, 0.01, 0.05, 0.1, 0.5, 0.8, 1.0\}$.
Detailed default hyper-parameters are summarized in~\cref{tab:app:hyperparam}.

\setlength{\tabcolsep}{8pt}
\begin{table}[h]
\centering
\caption{Architecture of ConvNet}
\resizebox{0.8\linewidth}{!}{
\begin{tabular}{llc}
\toprule
\noalign{\smallskip}
    \multirow{1}{*}{\begin{tabular}[c]{@{}c@{}} {\bftab Block} \end{tabular}} &
    \multirow{1}{*}{\begin{tabular}[c]{@{}c@{}} {\bftab Layers} \end{tabular}} &
    \begin{tabular}[c]{@{}c@{}} {\bftab Repetition} \end{tabular}
    \\          
\noalign{\smallskip}
\hline
\noalign{\smallskip}
 & \makecell[l]{Conv(3, 32, k=3, s=1), NormLayer(32), ReLU()} & 1\\
 & \makecell[l]{Conv(32, 64, k=3, s=2), NormLayer(64), ReLU()} & 1 \\
 & \makecell[l]{Conv(64, 64, k=3, s=2), NormLayer(64), ReLU()} & 1 \\
 & \makecell[l]{Conv(64, 64, k=3, s=1), NormLayer(64), ReLU()} & 1 \\
 & \makecell[l]{Conv(64, 128, k=3, s=2), NormLayer(128), ReLU()} & 1 \\
 & \makecell[l]{Conv(128, 128, k=3, s=1), NormLayer(128), ReLU()} & 1 \\
 & \makecell[l]{Conv(128, 256, k=3, s=2), NormLayer(256), ReLU()} & 1 \\
\noalign{\smallskip}
\hline
\noalign{\smallskip}
   & \makecell[l]{Flatten()} & 1 \\
   & FeatureNorm() \texttt{if use} \method & 1 \\
   & \makecell[l]{FC(1024, 100, bias=False)} & 1 \\
\bottomrule
\end{tabular}
}
\label{tab:model:convnet}
\end{table}
\setlength{\tabcolsep}{8pt}
\begin{table}[h]
\centering
\caption{Architecture of VGG13}
\resizebox{0.8\linewidth}{!}{
\begin{tabular}{llc}
\toprule
\noalign{\smallskip}
    \multirow{1}{*}{\begin{tabular}[c]{@{}c@{}} {\bftab Block} \end{tabular}} &
    \multirow{1}{*}{\begin{tabular}[c]{@{}c@{}} {\bftab Layers} \end{tabular}} &
    \begin{tabular}[c]{@{}c@{}} {\bftab Repetition} \end{tabular}
    \\          
\noalign{\smallskip}
\hline
\noalign{\smallskip}
 & \makecell[l]{Conv(3, 64, k=3, s=1, p=1), NormLayer(64), ReLU()} & 1\\
 & \makecell[l]{Conv(64, 64, k=3, s=1, p=1), NormLayer(64), ReLU()} & 1 \\
 & MaxPool2d(k=2, s=2) & 1 \\
 & \makecell[l]{Conv(64, 128, k=3, s=1, p=1), NormLayer(128), ReLU()} & 1 \\
 & \makecell[l]{Conv(128, 128, k=3, s=1, p=1), NormLayer(128), ReLU()} & 1 \\
  & MaxPool2d(k=2, s=2) & 1 \\
 & \makecell[l]{Conv(128, 256, k=3, s=1, p=1), NormLayer(256), ReLU()} & 1 \\
 & \makecell[l]{Conv(256, 256, k=3, s=1, p=1), NormLayer(256), ReLU()} & 1 \\
 & MaxPool2d(k=2, s=2) & 1 \\
  & \makecell[l]{Conv(256, 512, k=3, s=1, p=1), NormLayer(256), ReLU()} & 1 \\
 & \makecell[l]{Conv(512, 512, k=3, s=1, p=1), NormLayer(256), ReLU()} & 1 \\
 & MaxPool2d(k=2, s=2) & 1 \\
 & \makecell[l]{Conv(512, 512, k=3, s=1, p=1), NormLayer(256), ReLU()} & 2 \\
 & MaxPool2d(k=\textbf{TIN\_S}, s=\textbf{TIN\_S}) & 1 \\
\noalign{\smallskip}
\hline
\noalign{\smallskip}
   & AvgPool2d(k=1, s=1) & \\ 
   & \makecell[l]{Flatten()} & 1 \\
   & FeatureNorm() \texttt{if use} \method & 1 \\
   & \makecell[l]{FC(512, 100, bias=False)} & 1 \\
\bottomrule
\noalign{\smallskip}
\multicolumn{2}{l}{*: \textbf{TIN\_S}=1 if dataset is CIFAR-100 and \textbf{TIN\_S}=2 if dataset is TinyImageNet.}\\
\end{tabular}
}
\label{tab:model:vgg}
\end{table}
\setlength{\tabcolsep}{8pt}
\begin{table}[h]
\centering
\caption{Architecture of SENet18}
\resizebox{0.8\linewidth}{!}{
\begin{tabular}{llc}
\toprule
\noalign{\smallskip}
    \multirow{1}{*}{\begin{tabular}[c]{@{}c@{}} {\bftab Block} \end{tabular}} &
    \multirow{1}{*}{\begin{tabular}[c]{@{}c@{}} {\bftab Layers} \end{tabular}} &
    \begin{tabular}[c]{@{}c@{}} {\bftab Repetition} \end{tabular}
    \\          
\noalign{\smallskip}
\hline
\noalign{\smallskip}
  & \makecell[l]{Conv(3, 64, k=3, s=1, p=1), NormLayer(64), ReLU()} & 1\\
\cmidrule(l{5pt}r{5pt}){1-3}

\multirow{2}{*}[-1.5em]{B1} & \makecell[l]{Conv(64, 64, k=3, s=\textbf{TIN\_S}, p=1), NormLayer(64), ReLU()\\Conv(64, 64, k=3, s=1, p=1), NormLayer(64), ReLU()\\SquzzeExcitationModule()} & 1\\
\cmidrule(l{5pt}r{5pt}){2-3}
& \makecell[l]{Conv(64, 64, k=3, s=1, p=1), NormLayer(64), ReLU()\\Conv(64, 64, k=3, s=1, p=1), NormLayer(64), ReLU()\\SquzzeExcitationModule()} & 1\\
\cmidrule(l{5pt}r{5pt}){1-3}

\multirow{2}{*}[-1.5em]{B2} & \makecell[l]{Conv(64, 128, k=3, s=2, p=1), NormLayer(128), ReLU()\\Conv(128, 128, k=3, s=1, p=1), NormLayer(128), ReLU()\\SquzzeExcitationModule()} & 1\\
\cmidrule(l{5pt}r{5pt}){2-3}
& \makecell[l]{Conv(128, 128, k=3, s=1, p=1), NormLayer(128), ReLU()\\Conv(128, 128, k=3, s=1, p=1), NormLayer(128), ReLU()\\SquzzeExcitationModule()} & 1\\
\cmidrule(l{5pt}r{5pt}){1-3}

\multirow{2}{*}[-1.5em]{B3} & \makecell[l]{Conv(128, 256, k=3, s=2, p=1), NormLayer(256), ReLU()\\Conv(256, 256, k=3, s=1, p=1), NormLayer(256), ReLU()\\SquzzeExcitationModule()} & 1\\
\cmidrule(l{5pt}r{5pt}){2-3}
& \makecell[l]{Conv(256, 256, k=3, s=1, p=1), NormLayer(256), ReLU()\\Conv(256, 256, k=3, s=1, p=1), NormLayer(256), ReLU()\\SquzzeExcitationModule()} & 1\\
\cmidrule(l{5pt}r{5pt}){1-3}

\multirow{2}{*}[-1.5em]{B4} & \makecell[l]{Conv(256, 512, k=3, s=2, p=1), NormLayer(512), ReLU()\\Conv(512, 512, k=3, s=1, p=1), NormLayer(256), ReLU()\\SquzzeExcitationModule()} & 1\\
\cmidrule(l{5pt}r{5pt}){2-3}
& \makecell[l]{Conv(512, 512, k=3, s=1, p=1), NormLayer(512), ReLU()\\Conv(512, 512, k=3, s=1, p=1), NormLayer(512), ReLU()\\SquzzeExcitationModule()} & 1\\

\noalign{\smallskip}
\hline
\noalign{\smallskip}
   & AvgPool2d(k=4, s=4) & 1 \\
   & \makecell[l]{Flatten()} & 1 \\
   & FeatureNorm() \texttt{if use} \method & 1 \\
   & \makecell[l]{FC(512, 100, bias=False)} & 1 \\
\bottomrule
\noalign{\smallskip}
\multicolumn{2}{l}{*: \textbf{TIN\_S}=1 if dataset is CIFAR-100 and \textbf{TIN\_S}=2 if dataset is TinyImageNet.}\\
\end{tabular}
}
\label{tab:model:senet}
\end{table}
\setlength{\tabcolsep}{8pt}
\begin{table}[h]
\centering
\caption{Architecture of MobileNetV2.}
\resizebox{0.8\linewidth}{!}{
\begin{tabular}{llc}
\toprule
\noalign{\smallskip}
    \multirow{1}{*}{\begin{tabular}[c]{@{}c@{}} {\bftab Block} \end{tabular}} &
    \multirow{1}{*}{\begin{tabular}[c]{@{}c@{}} {\bftab Layers} \end{tabular}} &
    \begin{tabular}[c]{@{}c@{}} {\bftab Repetition} \end{tabular}
    \\          
\noalign{\smallskip}
\hline
\noalign{\smallskip}
  & \makecell[l]{Conv(3, 32, k=3, s=1), NormLayer(32), ReLU()} & 1\\
\cmidrule(l{5pt}r{5pt}){1-3}

B1 & \makecell[l]{Conv(32, 32, k=1, s=1), NormLayer(32), ReLU()\\Conv(32, 32, k=3, s=1, p=1, g=32), NormLayer(32), ReLU()\\Conv(32, 16, k=1, s=1), NormLayer(16), ReLU()} & 1\\
\cmidrule(l{5pt}r{5pt}){1-3}

\multirow{2}{*}[-1.5em]{B2} & \makecell[l]{Conv(16, 96, k=1, s=1), NormLayer(96), ReLU()\\Conv(96, 96, k=3, s=\textbf{TIN\_S}, p=1, g=96), NormLayer(96), ReLU()\\Conv(96, 24, k=1, s=1), NormLayer(24), ReLU()} & 1\\
\cmidrule(l{5pt}r{5pt}){2-3}
& \makecell[l]{Conv(24, 144, k=1, s=1), NormLayer(144), ReLU()\\Conv(144, 144, k=3, s=1, p=1, g=144), NormLayer(144), ReLU()\\Conv(144, 24, k=1, s=1), NormLayer(24), ReLU()} & 1\\
\cmidrule(l{5pt}r{5pt}){1-3}

\multirow{2}{*}[-1.5em]{B3} & \makecell[l]{Conv(24, 144, k=1, s=1), NormLayer(144), ReLU()\\Conv(144, 144, k=3, s=2, p=1, g=144), NormLayer(144), ReLU()\\Conv(144, 32, k=1, s=1), NormLayer(32), ReLU()} & 1\\
\cmidrule(l{5pt}r{5pt}){2-3}
& \makecell[l]{Conv(32, 192, k=1, s=1), NormLayer(192), ReLU()\\Conv(192, 192, k=3, s=1, p=1, g=192), NormLayer(192), ReLU()\\Conv(192, 32, k=1, s=1), NormLayer(32), ReLU()} & 2\\
\cmidrule(l{5pt}r{5pt}){1-3}

\multirow{2}{*}[-1.5em]{B4} & \makecell[l]{Conv(32, 192, k=1, s=1), NormLayer(192), ReLU()\\Conv(192, 192, k=3, s=2, p=1, g=192), NormLayer(192), ReLU()\\Conv(192, 64, k=1, s=1), NormLayer(64), ReLU()} & 1\\
\cmidrule(l{5pt}r{5pt}){2-3}
& \makecell[l]{Conv(64, 384, k=1, s=1), NormLayer(384), ReLU()\\Conv(384, 384, k=3, s=1, p=1, g=384), NormLayer(384), ReLU()\\Conv(384, 64, k=1, s=1), NormLayer(64), ReLU()} & 3\\
\cmidrule(l{5pt}r{5pt}){1-3}

\multirow{2}{*}[-1.5em]{B5} & \makecell[l]{Conv(64, 384, k=1, s=1), NormLayer(384), ReLU()\\Conv(384, 384, k=3, s=1, p=1, g=384), NormLayer(384), ReLU()\\Conv(384, 96, k=1, s=1), NormLayer(96), ReLU()} & 1\\
\cmidrule(l{5pt}r{5pt}){2-3}
& \makecell[l]{Conv(96, 576, k=1, s=1), NormLayer(576), ReLU()\\Conv(576, 576, k=3, s=1, p=1, g=576), NormLayer(576), ReLU()\\Conv(576, 96, k=1, s=1), NormLayer(96), ReLU()} & 2\\
\cmidrule(l{5pt}r{5pt}){1-3}

\multirow{2}{*}[-1.5em]{B6} & \makecell[l]{Conv(96, 576, k=1, s=1), NormLayer(576), ReLU()\\Conv(576, 576, k=3, s=2, p=1, g=576), NormLayer(576), ReLU()\\Conv(576, 160, k=1, s=1), NormLayer(160), ReLU()} & 1\\
\cmidrule(l{5pt}r{5pt}){2-3}
& \makecell[l]{Conv(160, 960, k=1, s=1), NormLayer(960), ReLU()\\Conv(960, 960, k=3, s=1, p=1, g=960), NormLayer(960), ReLU()\\Conv(960, 160, k=1, s=1), NormLayer(160), ReLU()} & 2\\
\cmidrule(l{5pt}r{5pt}){1-3}

B7 & \makecell[l]{Conv(160, 960, k=1, s=1), NormLayer(960), ReLU()\\Conv(960, 960, k=3, s=1, p=1, g=960), NormLayer(960), ReLU()\\Conv(960, 320, k=1, s=1), NormLayer(320), ReLU()} & 1\\
\cmidrule(l{5pt}r{5pt}){1-3}

& Conv(320, 1280, k=1, s=1), NormLayer(1280), ReLU() & 1\\
\noalign{\smallskip}
\hline
\noalign{\smallskip}
   & AvgPool2d(k=4, s=4) & 1 \\
   & \makecell[l]{Flatten()} & 1 \\
   & FeatureNorm() \texttt{if use} \method & 1 \\
   & \makecell[l]{FC(1280, 100, bias=False)} & 1 \\
\bottomrule
\noalign{\smallskip}
\multicolumn{2}{l}{*: \textbf{TIN\_S}=1 if dataset is CIFAR-100 and \textbf{TIN\_S}=2 if dataset is TinyImageNet.}\\
\end{tabular}
}
\label{tab:model:mbv2}
\end{table}
\setlength{\tabcolsep}{8pt}
\begin{table}[h]
\centering
\caption{Architecture of ResNet18}
\resizebox{0.8\linewidth}{!}{
\begin{tabular}{llc}
\toprule
\noalign{\smallskip}
    \multirow{1}{*}{\begin{tabular}[c]{@{}c@{}} {\bftab Block} \end{tabular}} &
    \multirow{1}{*}{\begin{tabular}[c]{@{}c@{}} {\bftab Layers} \end{tabular}} &
    \begin{tabular}[c]{@{}c@{}} {\bftab Repetition} \end{tabular}
    \\          
\noalign{\smallskip}
\hline
\noalign{\smallskip}
  & \makecell[l]{Conv(3, 64, k=3, s=1, p=1), NormLayer(64), ReLU()} & 1\\
\cmidrule(l{5pt}r{5pt}){1-3}

\multirow{2}{*}[-.9em]{B1} & \makecell[l]{Conv(64, 64, k=3, s=\textbf{TIN\_S}, p=1), NormLayer(64), ReLU()\\Conv(64, 64, k=3, s=1, p=1), NormLayer(64), ReLU()} & 1\\
\cmidrule(l{5pt}r{5pt}){2-3}
& \makecell[l]{Conv(64, 64, k=3, s=1, p=1), NormLayer(64), ReLU()\\Conv(64, 64, k=3, s=1, p=1), NormLayer(64), ReLU()} & 1\\
\cmidrule(l{5pt}r{5pt}){1-3}

\multirow{2}{*}[-.9em]{B2} & \makecell[l]{Conv(64, 128, k=3, s=2, p=1), NormLayer(128), ReLU()\\Conv(128, 128, k=3, s=1, p=1), NormLayer(128), ReLU()} & 1\\
\cmidrule(l{5pt}r{5pt}){2-3}
& \makecell[l]{Conv(128, 128, k=3, s=1, p=1), NormLayer(128), ReLU()\\Conv(128, 128, k=3, s=1, p=1), NormLayer(128), ReLU()} & 1\\
\cmidrule(l{5pt}r{5pt}){1-3}

\multirow{2}{*}[-.9em]{B3} & \makecell[l]{Conv(128, 256, k=3, s=2, p=1), NormLayer(256), ReLU()\\Conv(256, 256, k=3, s=1, p=1), NormLayer(256), ReLU()} & 1\\
\cmidrule(l{5pt}r{5pt}){2-3}
& \makecell[l]{Conv(256, 256, k=3, s=1, p=1), NormLayer(256), ReLU()\\Conv(256, 256, k=3, s=1, p=1), NormLayer(256), ReLU()} & 1\\
\cmidrule(l{5pt}r{5pt}){1-3}

\multirow{2}{*}[-.9em]{B4} & \makecell[l]{Conv(256, 512, k=3, s=2, p=1), NormLayer(512), ReLU()\\Conv(512, 512, k=3, s=1, p=1), NormLayer(256), ReLU()} & 1\\
\cmidrule(l{5pt}r{5pt}){2-3}
& \makecell[l]{Conv(512, 512, k=3, s=1, p=1), NormLayer(512), ReLU()\\Conv(512, 512, k=3, s=1, p=1), NormLayer(512), ReLU()} & 1\\

\noalign{\smallskip}
\hline
\noalign{\smallskip}
   & AvgPool2d(k=4, s=4) & 1 \\
   & \makecell[l]{Flatten()} & 1 \\
   & FeatureNorm() \texttt{if use} \method & 1 \\
   & \makecell[l]{FC(512, 100, bias=False)} & 1 \\
\bottomrule
\noalign{\smallskip}
\multicolumn{2}{l}{*: \textbf{TIN\_S}=1 if dataset is CIFAR-100 and \textbf{TIN\_S}=2 if dataset is TinyImageNet.}\\
\end{tabular}
}
\label{tab:model:resnet}
\end{table}
\setlength{\tabcolsep}{14pt}
\begin{table}[h]
\centering
\caption{Summary of default hyper-parameters.}
\resizebox{0.98\linewidth}{!}{
\begin{tabular}{ll|cccc}
\toprule
\noalign{\smallskip}
    \multirow{1}{*}{\begin{tabular}[c]{@{}c@{}} {\bftab Method} \end{tabular}} &
    \multirow{1}{*}{\begin{tabular}[c]{@{}c@{}} {\bftab Hyper-parameters} \end{tabular}} &
    \begin{tabular}[c]{@{}c@{}} {\bftab IID} \end{tabular} & 
     $\boldsymbol{\alpha=0.5}$ &
    $\boldsymbol{\alpha=0.1}$ &
    {\bftab TinyImageNet}
    \\

\noalign{\smallskip}
\hline
\noalign{\smallskip}
\multirow{7}{*}{\begin{tabular}[l]{@{}l@{}} FedAvg\\ (MobileNetV2) \end{tabular}} & Rounds & \multicolumn{4}{c}{100} \\
& Optimizer & \multicolumn{4}{c}{SGD}\\
& Weights decay & \multicolumn{4}{c}{0.00001}\\
& Momentum & \multicolumn{4}{c}{0.9}\\
& Local epochs & \multicolumn{4}{c}{10}\\
& Local batch size & \multicolumn{4}{c}{64}\\
& Learning rate & \multicolumn{4}{c}{0.1}\\

\noalign{\smallskip}
\cmidrule(l{12pt}r{12pt}){1-6}
\noalign{\smallskip}
\multirow{2}{*}{\plus\ {\scriptsize CCVR}} & \# virtual features per class~\cite{luo2021no} & 500 & 500 & 500 & 1000\\ 
& fine-tuning learning rate~\cite{luo2021no} & 0.00001 & 0.00001 & 0.00001 & 0.00001 \\

\noalign{\smallskip}
\cmidrule(l{12pt}r{12pt}){1-6}
\noalign{\smallskip}
\multirow{1}{*}{\plus\ {\scriptsize BABU}} & learning rate & 0.1 & 0.1 & 0.1 & 0.01\\ 

\noalign{\smallskip}
\cmidrule(l{12pt}r{12pt}){1-6}
\noalign{\smallskip}
\multirow{1}{*}{\plus\ Ours} & learning rate & 0.5 & 0.5 & 1.0 & 0.5\\

\noalign{\smallskip}
\hline
\hline
\noalign{\smallskip}
\multirow{8}{*}{\begin{tabular}[l]{@{}l@{}} FedProx\\ (ResNet18) \end{tabular}} & Rounds & \multicolumn{4}{c}{100} \\
& Optimizer & \multicolumn{4}{c}{SGD}\\
& Weights decay & \multicolumn{4}{c}{0.00001}\\
& Momentum & \multicolumn{4}{c}{0.9}\\
& Local epochs & \multicolumn{4}{c}{10}\\
& Local batch size & \multicolumn{4}{c}{64}\\
& Learning rate & \multicolumn{4}{c}{0.1}\\
& $\mu$ & \multicolumn{4}{c}{0.001}\\

\noalign{\smallskip}
\cmidrule(l{12pt}r{12pt}){1-6}
\noalign{\smallskip}
\multirow{2}{*}{\plus\ {\scriptsize CCVR}} & \# virtual features per class~\cite{luo2021no} & 500 & 500 & 500 & 1000\\ 
& fine-tuning learning rate~\cite{luo2021no} & 0.00001 & 0.00001 & 0.00001 & 0.00001 \\

\noalign{\smallskip}
\cmidrule(l{12pt}r{12pt}){1-6}
\noalign{\smallskip}
\multirow{2}{*}{\plus\ {\scriptsize BABU}} & learning rate & 0.1 & 0.1 & 0.1 & 0.1\\ 
& $\mu$ & 0.001 & 0.001 & 0.001 & 0.001 \\

\noalign{\smallskip}
\cmidrule(l{12pt}r{12pt}){1-6}
\noalign{\smallskip}
\multirow{2}{*}{\plus\ Ours} & learning rate & 0.5 & 0.5 & 0.5 & 0.5\\ 
& $\mu$ & 0.0001 & 0.0001 & 0.0001 & 0.001 \\

\noalign{\smallskip}
\hline
\hline
\noalign{\smallskip}
\multirow{7}{*}{\begin{tabular}[l]{@{}l@{}} FedNova\\ (VGG13) \end{tabular}} & Rounds & \multicolumn{4}{c}{100} \\
& Optimizer & \multicolumn{4}{c}{SGD}\\
& Weights decay & \multicolumn{4}{c}{0.00001}\\
& Momentum & \multicolumn{4}{c}{0.9}\\
& Local epochs & \multicolumn{4}{c}{10}\\
& Local batch size & \multicolumn{4}{c}{64}\\
& Learning rate & \multicolumn{4}{c}{0.01}\\

\noalign{\smallskip}
\cmidrule(l{12pt}r{12pt}){1-6}
\noalign{\smallskip}
\multirow{2}{*}{\plus\ {\scriptsize CCVR}} & \# virtual features per class~\cite{luo2021no} & 500 & 500 & 500 & 1000\\ 
& fine-tuning learning rate~\cite{luo2021no} & 0.00001 & 0.00001 & 0.00001 & 0.00001 \\

\noalign{\smallskip}
\cmidrule(l{12pt}r{12pt}){1-6}
\noalign{\smallskip}
\multirow{1}{*}{\plus\ {\scriptsize BABU}} & learning rate & 0.01 & 0.01 & 0.01 & 0.001\\ 

\noalign{\smallskip}
\cmidrule(l{12pt}r{12pt}){1-6}
\noalign{\smallskip}
\multirow{1}{*}{\plus\ Ours} & learning rate & 0.1 & 0.1 & 0.1 & 0.1\\

\noalign{\smallskip}
\hline
\hline
\noalign{\smallskip}
\multirow{10}{*}{\begin{tabular}[l]{@{}l@{}} FedOpt\\ (SENet18) \end{tabular}} & Rounds & \multicolumn{4}{c}{100} \\
& Optimizer & \multicolumn{4}{c}{SGD}\\
& Weights decay & \multicolumn{4}{c}{0.00001}\\
& Momentum & \multicolumn{4}{c}{0.9}\\
& Local epochs & \multicolumn{4}{c}{10}\\
& Local batch size & \multicolumn{4}{c}{64}\\
& Local learning rate & \multicolumn{4}{c}{0.01}\\
& On-server optimizer~\cite{reddi2020adaptive} & \multicolumn{4}{c}{SGD} \\
& On-server learning rate~\cite{reddi2020adaptive} & \multicolumn{4}{c}{1.0} \\
& On-server momentum~\cite{reddi2020adaptive} & \multicolumn{4}{c}{0.3} \\

\noalign{\smallskip}
\cmidrule(l{12pt}r{12pt}){1-6}
\noalign{\smallskip}
\multirow{2}{*}{\plus\ {\scriptsize CCVR}} & \# virtual features per class~\cite{luo2021no} & 500 & 500 & 500 & 1000\\ 
& fine-tuning learning rate~\cite{luo2021no} & 0.00001 & 0.00001 & 0.00001 & 0.00001 \\

\noalign{\smallskip}
\cmidrule(l{12pt}r{12pt}){1-6}
\noalign{\smallskip}
\multirow{1}{*}{\plus\ {\scriptsize BABU}} & Local learning rate & 0.01 & 0.01 & 0.01 & 0.01\\ 

\noalign{\smallskip}
\cmidrule(l{12pt}r{12pt}){1-6}
\noalign{\smallskip}
\multirow{1}{*}{\plus\ Ours} & Local learning rate & 0.5 & 0.5 & 0.5 & 0.5\\ 

\bottomrule
\end{tabular}
}
\label{tab:app:hyperparam}
\end{table}

\end{document}